\documentclass[conference]{IEEEtran}
\IEEEoverridecommandlockouts

\usepackage{cite}
\usepackage{amsmath,amssymb,amsfonts}
\usepackage{graphicx}
\usepackage{textcomp}
\usepackage{xcolor}
\usepackage[font=small]{caption}

\usepackage{mathtools}
\usepackage{enumitem}
\usepackage{wrapfig}
\usepackage{booktabs}

\usepackage{algorithm}
\usepackage[noend]{algorithmic}
\usepackage{multicol}

\makeatletter
\newcommand\notsotiny{\@setfontsize\notsotiny\@vipt\@viipt}
\makeatother

\usepackage{hyperref}
\PassOptionsToPackage{hyphens}{url}\usepackage{hyperref}
\usepackage{cleveref}

\def\BibTeX{{\rm B\kern-.05em{\sc i\kern-.025em b}\kern-.08em
    T\kern-.1667em\lower.7ex\hbox{E}\kern-.125emX}}

\usepackage{amsmath}

\ifCLASSOPTIONcompsoc
    \usepackage[caption=false,font=normalsize,labelfont=sf,textfont=sf]{subfig}
\else
    \usepackage[caption=false,font=footnotesize]{subfig}
\fi

\begin{document}




\title{Accelerating Federated Learning in Heterogeneous Data and Computational Environments}



\author{\IEEEauthorblockN{Dimitris Stripelis}
\IEEEauthorblockA{\textit{Information Sciences Institute} \\
\textit{University of Southern California}\\
Los Angeles, CA, 90089 \\
stripeli@isi.edu}
\and
\IEEEauthorblockN{Jos\'{e} Luis Ambite}
\IEEEauthorblockA{\textit{Information Sciences Institute} \\
\textit{University of Southern California}\\
Los Angeles, CA, 90089 \\
ambite@isi.edu}
}


\maketitle

\begin{abstract}
There are situations where data relevant to a machine learning problem are distributed among multiple locations that cannot share the data due to regulatory, competitiveness, or privacy reasons. For example, data present in users' cellphones, manufacturing data of companies in a given industrial sector, or medical records located at different hospitals. Moreover, participating sites often have different data distributions and computational capabilities. Federated Learning provides an approach to learn a joint model over all the available data in these environments. In this paper, we introduce a novel distributed validation weighting scheme (DVW), which evaluates the performance of a learner in the federation against a distributed validation set. Each learner reserves a small portion (e.g., 5\%) of its local training examples as a validation dataset and allows other learners models to be evaluated against it. We empirically show that DVW results in better performance compared to established methods, such as FedAvg, both under \textit{synchronous} and \textit{asynchronous} communication protocols in \textit{data and computationally heterogeneous} environments.
\end{abstract}

\begin{IEEEkeywords}
Federated Learning; Neural Networks; Heterogeneous Computing Environments; Heterogeneous Data
\end{IEEEkeywords}

\section{Introduction}
Data useful for a machine learning problem is often generated at multiple, distributed locations. In many situations these data cannot be exported from their original location due to regulatory, competitiveness, or privacy reasons. A primary motivating example is health records, which are heavily regulated and protected, restricting the ability to analyze large datasets. Industrial data (e.g., accident or safety data) is also not shared due to competitiveness reasons. Additionally, given recent high-profile data leak incidents, e.g., Facebook in 2018, more strict data ownership laws have been enacted in many countries, such as the European Union's General Data Protection Regulation (GDPR), China's Cyber Security Law and General Principles of Civil Law, and the California Consumer Privacy Act (CCPA). 

These situations bring data distribution, security, and privacy to the forefront of machine learning and impose new challenges on how data should be processed and analyzed. A recent promising solution is Federated Learning (FL) \cite{mcmahan2017communication,konevcny2016federated,yang2019federated}, which can learn deep neural networks from data silos \cite{jain2003out}, providing privacy and security guarantees \cite{bonawitz2017practical}, by collaboratively training models that aggregate locally-computed updates (e.g., gradients) under a centralized (e.g., central parameter server) or a decentralized (e.g., peer-to-peer) learning topology \cite{li2020federated}. 


Most of the existing FL work focuses on synchronous communication protocols \cite{mcmahan2017communication,smith2017federated,bonawitz2019towards} over a federation of learners with homogeneous computational capabilities and similar data distributions per learner.
In this paper, we study Federated Learning under both \textit{synchronous} and \textit{asynchronous} parameter-update protocols in \textit{homogeneous} and \textit{heterogeneous} computing environments over varying data amounts and assignments of IID and/or non-IID examples across learners. 


We propose a novel Federated Learning training architecture that is resilient to system and statistical heterogeneity by combining the learners' locally-trained models based on their performance against a stratified distributed validation dataset. Our contributions are:

\begin{itemize}
    \item[$\bullet$] A Federated Learning framework, \textit{Metis}, designed to explore, modularly, distributed learning protocols, under synchronous and asynchronous communication, in homogeneous and heterogeneous environments. 
    \item[$\bullet$] A novel Federated Learning training scheme, called Distributed Validation Weighting (DVW), which reserves a small validation dataset at every learner's local site to evaluate other learners' models and provide an objective value of the model performance in the federation, which yields more effective model mixing.  
    \item[$\bullet$] A community computation approach for asynchronous protocols that always considers the most recently committed model of every learner and computes the community/federation model in constant time.
    \item[$\bullet$] An asynchronous communication protocol with an adaptive update frequency on a per learner basis.
    \item[$\bullet$] A systematic empirical evaluation of Federated Learning training schemes, showing faster convergence of our proposed asynchronous DVW scheme and better generalization on highly diverse data distributions.
\end{itemize}

\section{Federated Learning: \\ Background and Related Work}

Federated Learning over neural networks was introduced by McMahan et al.~\cite{mcmahan2017communication} for user data in mobile phones. Their algorithm, Federated Average (FedAvg), follows a synchronous communication protocol, where each learner (phone) trains a neural network for a fixed number of epochs on its local dataset. Once all learners (or a subset) finish their assigned training, the system computes a community model that is a weighted average of each of the learners' local models, with the weight of each learner in the federation based on the number of its local training examples. The new community model is then distributed to all learners and the process repeats. This approach, which we call \textit{SyncFedAvg}, has catalyzed much recent work~\cite{smith2017federated,bonawitz2019towards,li2018federated}. 




\textbf{Distributed Stochastic Gradient Descent (SGD).} The synchronous and asynchronous FL settings that we investigate are closely related to stochastic optimization in distributed and parallel systems \cite{bertsekas1983distributed,bertsekas1989parallel}, as well as in synchronous distributed SGD optimization \cite{chen2016revisiting}. The problem of delayed (i.e., stale) gradient updates due to asynchronicity is well known \cite{lian2015asynchronous,recht2011hogwild,agarwal2011distributed}, with \cite{lian2015asynchronous} providing theoretical support for nonconvex optimization functions under the IID assumption. We study FL in more general, Non-IID settings. 


\textbf{FL SGD.} In heterogeneous FL, learners can drift too far away from the global optimal model. An approach to tackle drift is to decouple the SGD optimization into local (learner side) and global (server side) \cite{reddi2020adaptive,hsu2019measuring}.
In \cite{reddi2020adaptive}, after computing the weighted average of the clients updated weights ("pseudo-gradients"), a new community model is computed through adaptive SGD optimizers that target to optimize the global objective.  
In \cite{hsu2019measuring}, the authors investigate a momentum-based update rule between the previous community model and the newly computed weighted average of the clients models.
Other approaches \cite{li2018federated,xie2019asynchronous} directly addresses drift by introducing a regularization term in the clients local objective, which accounts for the divergence of the local solution from the global solution.
In our work, we use Momentum SGD, which shows accelerated convergence in FL compared to Vanilla SGD \cite{liu2020accelerating}, and we define a new mixing strategy of the clients local models in the server side based on their performance on a distributed validation dataset (cf. Section~\ref{sec:DVWScheme}).

\textbf{FL Convergence.} Convergence guarantees for computational environments with heterogeneous resources have been studied in \cite{li2018federated,wang2019adaptive}. 
FedProx \cite{li2018federated} studied the convergence rate of FedAvg over $B$-dissimilar learners local solutions ($B=1$ IID distributions, $B>1$ non-IID distributions).
Wang et al. \cite{wang2019adaptive} studied adaptive FL in mobile edge computing environments under resource budget constraints with arbitrary local updates between clients.
Li et al. \cite{Li2020On} provide convergence guarantees over full and partial device participation for FedAvg. 
FedAsync \cite{xie2019asynchronous} provides convergence guarantees for asynchronous environments and a community model that is a weighted average of local models based on staleness. 
In our work, we empirically study the convergence of our DVW weighting scheme over synchronous and asynchronous heterogeneous environments with dissimilar data distributions and full client participation in the datacenter setting.



\textbf{FL Privacy.} Privacy is a critical challenge in FL. Although we do not address privacy in this paper, our framework can be extended to incorporate standard privacy-aware techniques such as differential privacy \cite{abadi2016deep,mcmahan2017learning}, secure multi-party computation (MPC) \cite{bonawitz2017practical, mohassel2017secureml, kilbertus2018blind}, homomorphic Encryption\cite{rivest1978data}, or Paillier partial homomorphic encryption scheme~\cite{paillier-crypto}.

\textbf{Federated Optimization.} In Federated Learning the goal is to find the optimal set of parameters $w^*$ that minimize the global objective function $f(w)$:
\begin{equation}\label{eq:FederatedFunction}
w^*=\underset{w}{\mathrm{argmin}} f(w) \quad\text{where}\quad f(w)=\sum_{k=1}^{N}\frac{p_k}{\mathcal{P}}F_k(w)
\end{equation}
where N denotes the number of participating learners, $p_k$ the contribution of learner $k$ in the federation,
$\mathcal{P}=\sum p_k$ the normalization factor (thus, $\sum_{k}^N \frac{p_k}{\mathcal{P}}=1$), and $F_k(w)$ the local objective function of learner $k$. 
We refer to the model computed using Equation \ref{eq:FederatedFunction} as the community model $w_c$.
Every learner computes its local objective by minimizing the local empirical risk over its training set $D_k^T$ as $F_k(w) = \mathbb{E}_{x_k \sim D_k^T}[\ell_k(w;x_k)]$, with $\ell_k$ being the loss function. 
For example, in the FedAvg weighting scheme, the contribution value for any learner $k$ is equal to its local training set size, $p_k = \left|D_k\right|$ and $\mathcal{P}=\sum\frac{\left|D_k^T\right|}{\left|D^T\right|}$, where $D^T=\bigcup_{k}^N D_k^T$ and $\left|D^T\right|=\sum_{k}^N \left|D_k^T\right|$. 
%
%
The contribution value $p_k$ can be static, or dynamically defined at run time (cf. Section~\ref{sec:StoppingCriteriaAsyncDVW}).


We use Stochastic Gradient Descent with Momentum as a learner's local objective solver\cite{liu2020accelerating} (as opposed to SGD in  \cite{mcmahan2017communication}) with the local solution $w_{t+1}$ at iteration $t$ computed as:
\begin{equation}\label{eq:LocalSGDWithMomentum}
    \begin{gathered}
        u_{t+1} \leftarrow \gamma u_{t} + \nabla F_k(w_t) \\
        w_{t+1} \leftarrow w_{t} - \eta u_{t+1}
    \end{gathered}    
\end{equation}
%
with $\eta$ denoting the learning rate, $u$ the momentum term and $\gamma$ the momentum attenuation factor. FedProx\cite{li2018federated} is a variation of the local SGD solver that introduces a proximal term in the update rule to regularize the local updates based on the divergence of the local solution from the global solution.
The Fedprox regularization term is also used in FedAsync \cite{xie2019asynchronous}, which we empirically compare to in section \ref{sec:Evaluation}.



\section{Metis Federated Learning Framework}\label{sec:MetisFederatedLearningFramework}

\begin{figure*}[htbp]
    \includegraphics[scale=0.67]{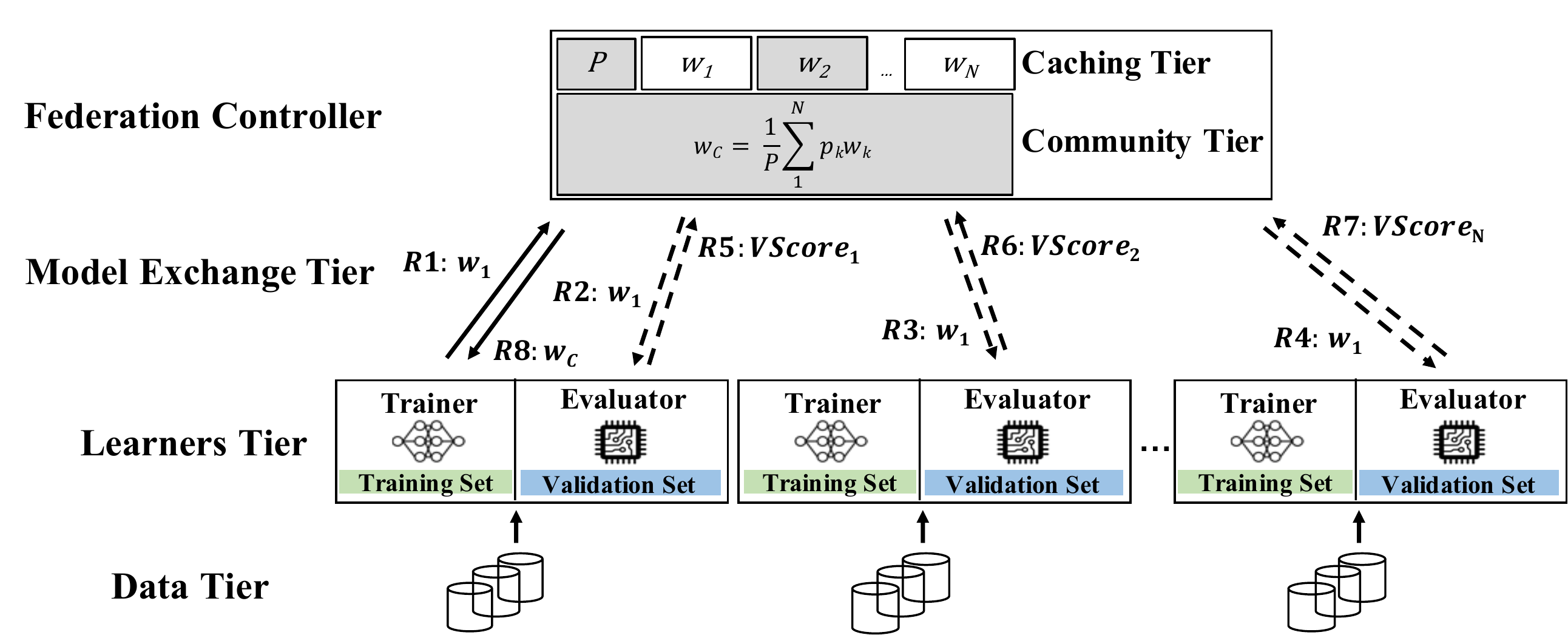}
    \captionsetup{justification=centering}
    \caption{Metis System Architecture}
    \label{fig:MetisSystemArchitecture}
\end{figure*}

We have designed a flexible Federated Learning architecture, called Metis, to explore different communication protocols and model aggregation weighting schemes (see Figure~\ref{fig:MetisSystemArchitecture}). 

\textbf{Federation Controller}. The centralized controller is a multi-threaded process with a modular 
design that integrates a collection of extensible microservices (i.e., caching and community tiers). The controller orchestrates the execution of the entire federation and is responsible to initiate the system pipeline, broadcast the initial community model and handle community model update requests.
The controller handles every incoming community update in a FIFO ordering through a mutual exclusive lock, ensuring system state linearizability \cite{herlihy1990linearizability}. Essentially, the federation controller is a materialized version of the Parameter Server \cite{abadi2016tensorflow,dean2012large} concept, widely used in distributed learning applications. 

\textbf{Community \& Caching Tier.} The community tier computes a new community model $w_c$ (Equation \ref{eq:FederatedFunction}), as a weighted average of the most recent model that each individual learner has committed to the controller. To facilitate this computation, it is natural to store in-memory or in disk the most recently received local model of every learner in the federation. Therefore, the memory and storage requirements of a community model depend on the number of local models contributing to the community model. For a synchronous protocol we always need to perform a pass over the entire collection of stored local models, with a computational cost $O(MN)$, where $M$ is the size of the model and $N$ is the number of learners. For an asynchronous protocol where update requests are generated at different paces, such a complete pass is redundant and we can leverage the existing cached/stored local models to compute a new community model in time $O(M)$, independent of the number of local models.

Some asynchronous community mixing approaches \cite{xie2019asynchronous} compute a weighted average using a mixing hyperparameter between the current community and the committing model of a requesting learner. In contrast, our DVW approach automatically assigns a unique weighting value to every contributed local model based on the learner performance in the federation, with no hyperparameter dependence (cf. Section \ref{sec:DVWScheme}).



The Caching Tier efficiently computes community models. Consider an unnormalized community model consisting of $m$ matrices, $W_c=\langle W_{c_1}, W_{c_2}, \ldots, W_{c_m}\rangle$, and a community normalization weighting factor $\mathcal{P}=\sum_{k=1}^{N} p_{k}$. 
Given a new request from learner $k$, with community contribution value $p_k$, the new normalization value is equal to, $\mathcal{P} = \mathcal{P} + p_k - p_k^{\prime}$, where $p_k^{\prime}$ is the learner's existing contribution value.
For every component matrix $W_{c_i}$ 
of the community model, the updated matrix is $W_{c_i} = W_{c_i} + p_k w_{k,i} - p_k^{\prime}w_{k,i}^{\prime}$, where $w_{k,i}, w_{k,i}^{\prime}$ are the new and existing component matrices for learner $k$. The new community model is $w_c = \frac{1}{\mathcal{P}} W_c$.

%

\begin{figure}
    \centering
    \includegraphics[width=\linewidth]{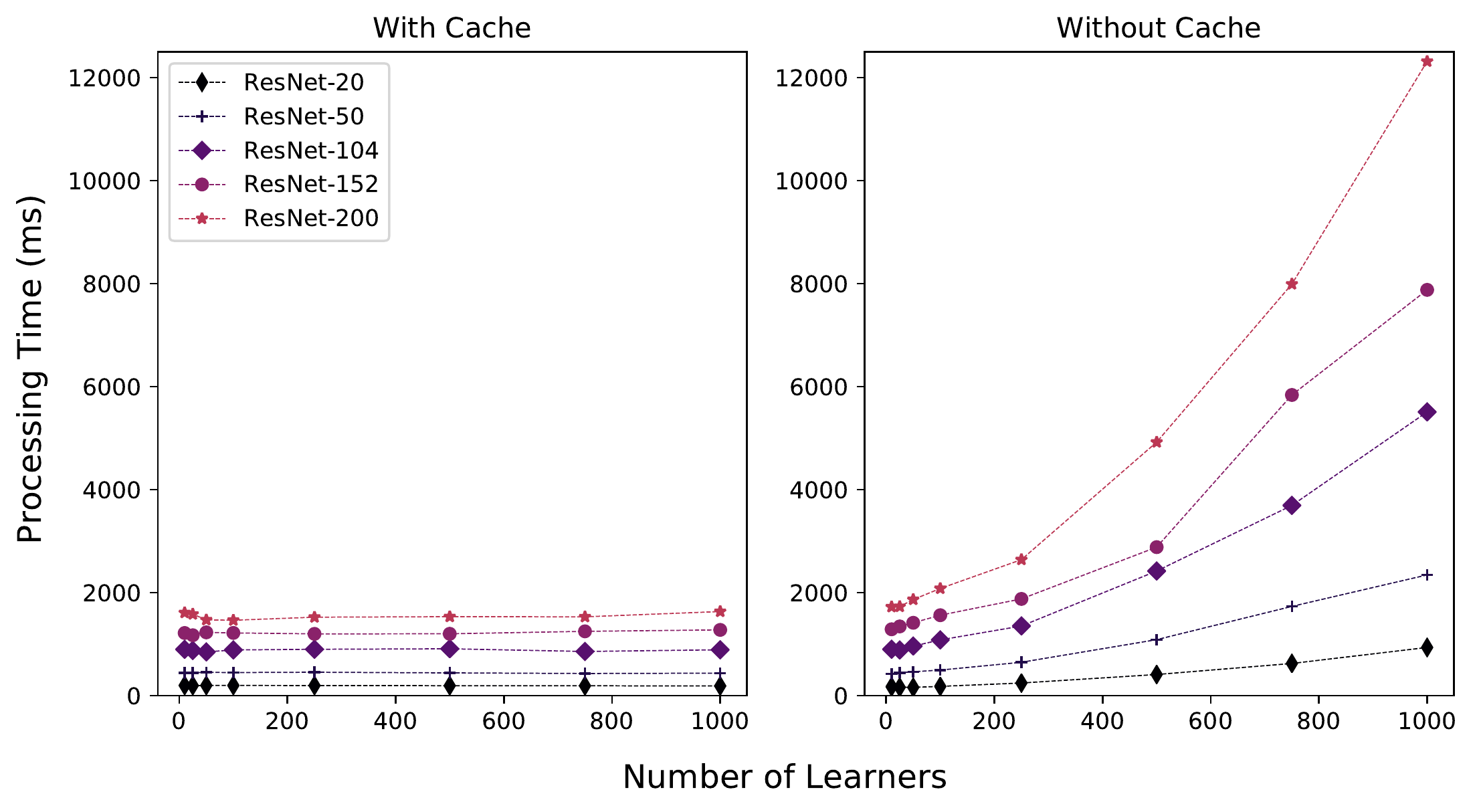}
    \caption{Community computation with (left) and without (right) cache}
    \label{fig:CommunityWithAndWithoutCache}
    \vspace{-5mm}
\end{figure}


Figure~\ref{fig:CommunityWithAndWithoutCache} shows the computation cost for different sizes of a ResNet community model (in the Cifar100 domain) as the federation increases to 1000 learners. With our caching mechanism the time remains constant, while it significantly increases without it. 

\textbf{Model Exchange Tier.} Every learner in the federation contacts the controller for a model update and sends its local model to the controller once it has finished training on its local training dataset. Upon the computation of the community model, the learner pulls the model from the controller and continues on its next training cycle. This exchange is represented with requests R1 and R8 in Figure \ref{fig:MetisSystemArchitecture}. 

\textbf{Learners \& Data Tier.} All learners train on the same neural network architecture with identical hyperparameter values (learning rate, batch size, etc.), starting from the same initial (random) model state, and using the same local SGD optimizer. 
The number of local epochs a learner performs before issuing an update request can be predefined or computed adaptively (cf. Section~\ref{sec:StoppingCriteriaAsyncDVW}). We will refer to local epochs and update frequency $uf$ interchangeably. Every learner trains on its own local dataset. No data is shared across learners. However, a learner may evaluate the model of another learner over its local validation set (cf. Section~\ref{sec:DVWScheme}).



\begin{algorithm*}[htpb]
    \caption{\texttt{Metis Framework with DVW.} Community model $w_c$ consisting of $m$ matrices is computed with $N$ learners, each indexed by $k$; $\gamma$ is the momentum attenuation factor; $\eta$ is the learning rate; $\beta$ is the batch size.}
    \label{alg:METIS}
    \vspace{-\baselineskip}
    \begin{multicols}{2}
        \begin{algorithmic}
        \renewcommand{\algorithmicrequire}{\textbf{Initialization}: $w_c, \gamma, \eta, \beta$}
        \REQUIRE
        \STATE
        \renewcommand{\algorithmicrequire}{\textbf{\underline{Synchronous}}}
        \REQUIRE
        \FOR{$t = 0, \dots, T-1$}
             \FOR{each client $k \in N$ \textbf{in parallel}}
                \STATE $w_k = \textsc{ClientOpt}(w_c, meta)$
                \STATE $p_k = 
                    \begin{cases} 
                        D_K^T & \text{\small if SyncFedAvg} \\ 
                        \textsc{Eval}(w_k) & \text{\small if DVW} \\ 
                    \end{cases}$
             \ENDFOR
             \STATE {$w_c = \sum_{k=1}^{N}\frac{p_{k}}{\mathcal{P}}w_k$ with $\mathcal{P}=\sum_{k}^N p_k$}
        \ENDFOR
        \STATE
        \renewcommand{\algorithmicrequire}{\textbf{\underline{Asynchronous}}}
        \REQUIRE
        \STATE $P=0$; $\forall k \in N, p_k=0$; $\forall i \in m, W_{c,i}=0$
        \STATE $\forall k \in N$ \textsc{ClientOpt}($w_c, meta$)
        \WHILE{\textbf{true}}
            \IF{(update request from client $k$ with model $w_k$)}
                \STATE $p_k = 
                    \begin{cases} 
                        D_K^T & \text{\small if AsyncFedAvg} \\ 
                        \textsc{Eval}(w_k) & \text{\small if DVW} \\ 
                    \end{cases}$
                \STATE $\mathcal{P} = \mathcal{P} + p_k - p_k^{\prime}$
                \FOR{$i \in m$}
                    \STATE $W_{c,i} = W_{c,i} + p_kw_{k,i} - p_k^{\prime}w_{k,i}^{\prime}$
                \ENDFOR    
                \STATE $w_c = \frac{1}{\mathcal{P}} W_c$
                \STATE Reply $w_c$ to client $k$
            \ENDIF
        \ENDWHILE
        \columnbreak
        \renewcommand{\algorithmicrequire}{\textbf{\textsc{ClientOpt($w_t, meta$):}}}
        \REQUIRE        
            \WHILE{\textbf{true}}
                \STATE $\mathcal{B} \leftarrow$ Split training data $D_k^{T}$ into batches of size $\beta$
                \FOR{$b \in \mathcal{B}$}
                    \STATE {$u_{t+1} = \gamma u_{t} - \eta\nabla F_k(w_t;b)$}
                    \STATE {$w_{t+1} = w_{t} + u_{t+1}$}
                \ENDFOR
                \STATE TriggerUpdate = 
                    $\begin{cases} 
                        \text{\textbf{sync/async (nonadaptive):}} \\
                            \ \textit{CurrentEpochs} > meta[\mathtt{uf}] \\
                        \text{\textbf{async with \textsc{Dvw} (adaptive):}} \\
                            \ \text{\textbf{(C1)}} \ Vpct \geq 0 \ \textit{OR}  \\
                            \ \text{\textbf{(C2)}} \ Vpct < 0 \ \text{\&\&} \\
                            \ \qquad |Vpct| \leq meta[\mathtt{VC_{Loss}}] \ \textit{OR} \\
                            \ \text{\textbf{(C3)}} \ \textit{Staleness} > \textit{Median(Staleness)}
                    \end{cases}$                    
                \IF{(TriggerUpdate)}
                    \STATE Send $w_{t+1}$ to controller
                \ENDIF
            \ENDWHILE 
        \STATE
        \renewcommand{\algorithmicrequire}{\textbf{\textsc{Eval($w$):}}}
        \REQUIRE
            \STATE $CM = 0_{C,C}$  \COMMENT{Cumulative confusion matrix of size CxC}
            \FOR{each client $k \in N$ \textbf{in parallel}}
                \STATE $CM = CM + \textsc{Evaluator}_{k}(w)$
            \ENDFOR
            \STATE $\textsc{Dvw} = {F_{1}^{mi}}(CM)$ \COMMENT{Equation \ref{eq:DVWMicroF1}}
            \STATE Return $\textsc{Dvw}$
    \end{algorithmic}
    
    \end{multicols}
    \vspace{-\baselineskip}
\end{algorithm*}

\textbf{Execution Pipeline.} Algorithm \ref{alg:METIS} describes the execution pipeline of the Metis framework for both synchronous and asynchronous communication protocols. In synchronous protocols, the controller waits for all the participating learners to finish training on their local training dataset before computing a community model, distributing it to the learners, and proceeding to the next global iteration.
In asynchronous protocols, the controller computes a community model whenever a single learner finishes its local training, using the caching mechanism, and sends it to that learner. 
In both cases, the controller assigns a contribution value $p_k$ to the local model $w_k$ that a learner $k$ shares with the community. For synchronous and asynchronous FedAvg (SyncFedAvg and AsyncFedAvg), this value is statically defined and based on the size of the learner's local training dataset, $D_k^T$. For other weighting schemes, e.g. DVW, the \textsc{Eval} procedure computes it dynamically. 

In the DVW scheme, the \textsc{Eval} procedure is responsible to evaluate a client's $k$ local model over the local validation dataset of every participating learner through its \textsc{Evaluator} service. Using the evaluation results from all the learners, the controller computes a unique contribution value for that local model. Since Metis is modular, it can support different weighting schemes and communication protocols. 



The \textsc{ClientOpt} implements the local training of each learner. Each learner requests a community update whenever a \textit{TriggerUpdate} condition is satisfied. For a non-adaptive execution, the total number of local iterations is a user-defined parameter. For adaptive execution, we show in Algorithm~\ref{alg:METIS} the conditions of the DVW scheme (cf. Section~\ref{sec:DVWScheme}). All the hyperparameters necessary to control the number of local iterations a learner needs to perform are passed through the metadata, $meta$, collection.

\section{Distributed Validation Weighting (DVW)} \label{sec:DVWScheme}

Our core goal is to define an objective metric to determine the weighting value of the local model of a learner in a heterogeneous federated learning environment, where each learner may have very different data distributions or computational capabilities. 
Instead of using a proxy metric, such as the size of the local training dataset, we measure the quality of the local model directly by constructing a distributed validation dataset and evaluate its performance against it. 

\textbf{Distributed Validation Dataset.} In our DVW approach, each learner $k$ reserves a small portion of its local training dataset for validation ($D_{k}^{V}$) and allows for other learners models to be evaluated over it. 
In our architecture, each learner supports an \textit{evaluator} service to test the performance of received models over its local validation dataset, as shown by requests R2, R3, and R4 in the \textit{Learners Tier} of Figure~\ref{fig:MetisSystemArchitecture}. 

Due to the intrinsic heterogeneity in the distribution and the size of the learners local datasets, we argue that every validation dataset in the federation needs to be a locally \textit{stratified} set of data samples in order to provide a true representation of the underlying data distribution of every learner.\footnote{Empirically, we observed unstable generalization with randomly generated validation datasets.}
Even though no local validation data sample ever leaves its original site, conceptually, the global validation dataset is created as if all the stratified validation datasets in the federation where grouped together. That is, the distributed validation dataset is $D_{F}^{V} = \bigcup_{k}^{N} D_{k}^{V}$ for a federation of $N$ learners. Even if each local validation dataset is small (e.g., 5\% of local training data), $D_{F}^{V}$ represents a large distributed validation dataset as a whole.
Given that the federation validation dataset is a consolidation of stratified data samples across all learners, it is a close representation of the federation domain for any combination of IID or non-IID data distributions and heterogeneous data sizes.



\textbf{Distributed Validation Weighting Schemes.} Upon a new community update request, the controller sends the requesting learner's local model to the evaluator service of all the rest of the learners, and retrieves and combines the validation quality metrics (requests R5, R6, R7 in Figure~\ref{fig:MetisSystemArchitecture}) to determine the weight of the learner in the federation. 
For classification tasks, as those presented in this work, these metrics are the confusion matrices, $CM$ of size $C \times C$ with $C$ referring to the number of classes, generated by each evaluator service over its validation dataset (see \textsc{Eval} procedure in Alg. \ref{alg:METIS}). 
For other tasks, such as deep regression, our approach can be extended to support additional evaluation metrics.

Our architecture can use a variety of performance metrics to assign a weight to a learner's model. 
%
%
Originally, we used accuracy as performance metric, but it is not robust to classification errors and tends to undervalue how well individual models (classifiers) perform across classes. Since we need a performance metric resilient to both balanced and imbalanced domains, we use the F1-measure as the classification performance metric in the experiments in Section~\ref{sec:Evaluation}.
Other classification quality assessment methods \cite{tharwat2018classification}, such as geometric mean, or the Mathews correlation coefficient, can also be used to implement the DVW scheme (cf. Section \ref{sec:Discussion}). 


There are several methods to compute the F-measure for multi-class settings, but the Micro F\textsubscript{1}-Score, $F_{1}^{mi}=\frac{2*TP}{2*TP + FP + FN}$ (where $TP, FP$ and $FN$ are the total number of true positives, false positives and false negatives across all classes) is an unbiased method that can handle high degrees of class imbalance \cite{forman2010apples}, and hence it is the metric we use in the DVW scheme.
To compute the DVW Micro F\textsubscript{1}-Score, the controller pools all the counts from the confusion matrix of each learner, and computes the cumulative number of true positives, false positives, and false negatives, for all the classes $C$ across all local validation datasets $V$ that constitute the global federation validation dataset. With the rows of the confusion matrix being the actual labels and its columns the prediction labels, we have:
\begin{equation*}
    \begin{aligned}
        \begin{pmatrix} 
            TP_\mathrm{C} = \sum_\mathrm{C}\sum_\mathrm{V} tp_c^\mathrm{v} = \sum_i CM[i,i] \\ 
            FP_\mathrm{C} = \sum_\mathrm{C}\sum_\mathrm{V} fp_c^\mathrm{v} = \sum_j(\sum_i CM[i,j]) - CM[j,j]\\
            FN_\mathrm{C} = \sum_\mathrm{C}\sum_\mathrm{V} fn_c^\mathrm{v} = \sum_i(\sum_j CM[i,j]) - CM[i,i]
        \end{pmatrix}
    \end{aligned}
\end{equation*}
And the final DVW Micro F\textsubscript{1}-Score is equal to:
\begin{equation}\label{eq:DVWMicroF1}
    DVW_{F_{1}^{mi}}=\frac{2* TP_\mathrm{C}}{2*TP_\mathrm{C} + FP_\mathrm{C} + FN_\mathrm{C}}
\end{equation}

\section{Adaptive Asynchronous DVW}
\label{sec:StoppingCriteriaAsyncDVW}

While existing work in synchronous and asynchronous Federated Learning defines a fixed update frequency for all learners \cite{mcmahan2017communication,xie2019asynchronous}, we propose an \textit{adaptive} mechanism based on the performance of each learner on its local validation dataset. Consequently, the importance of the local validation dataset is two-fold: first by serving as a testbed for evaluating federation models and second as a proxy to learn and adjust the update frequency of each learner. Intuitively, a learner requests an update when its local validation loss is not improving significantly or when its local model is becoming too stale. 

\textbf{Validation Loss Criterion.} To ensure that each learner is making good progress, we keep track of the percentage difference of the loss (cross entropy loss in our experiments) between two consecutive epochs of the learner's local model on its local validation set, i.e., $Vpct = 100 * (VLoss_{i} - VLoss_{i-1}) / {VLoss_{i-1}}$.
Since we do not want the learner to overfit its local dataset, the learner halts training and requests a community update when one of two conditions are met:

\textbf{(C1)} $Vpct \geq 0 $, or  
\textbf{(C2)} $Vpct < 0 \, \text{and} \, |Vpct| \leq VC_{Loss}$


\noindent Condition C1 indicates that the validation loss has increased or plateaued, and condition C2 captures the magnitude of the decrease. The term $VC_{Loss}$ is a user-defined threshold that signals when the improvements are too small and the learner may be better off asking for a new community model. These conditions can be seen as a form of early stopping \cite{prechelt1998early}. We refer to the above validation evaluation phase as a \textit{Validation Cycle} ($VC$) with length equal to the total number of local epochs performed by the learner. 


\begin{figure}
    \vspace*{-3mm}
    \centering
    \includegraphics[width=\linewidth]{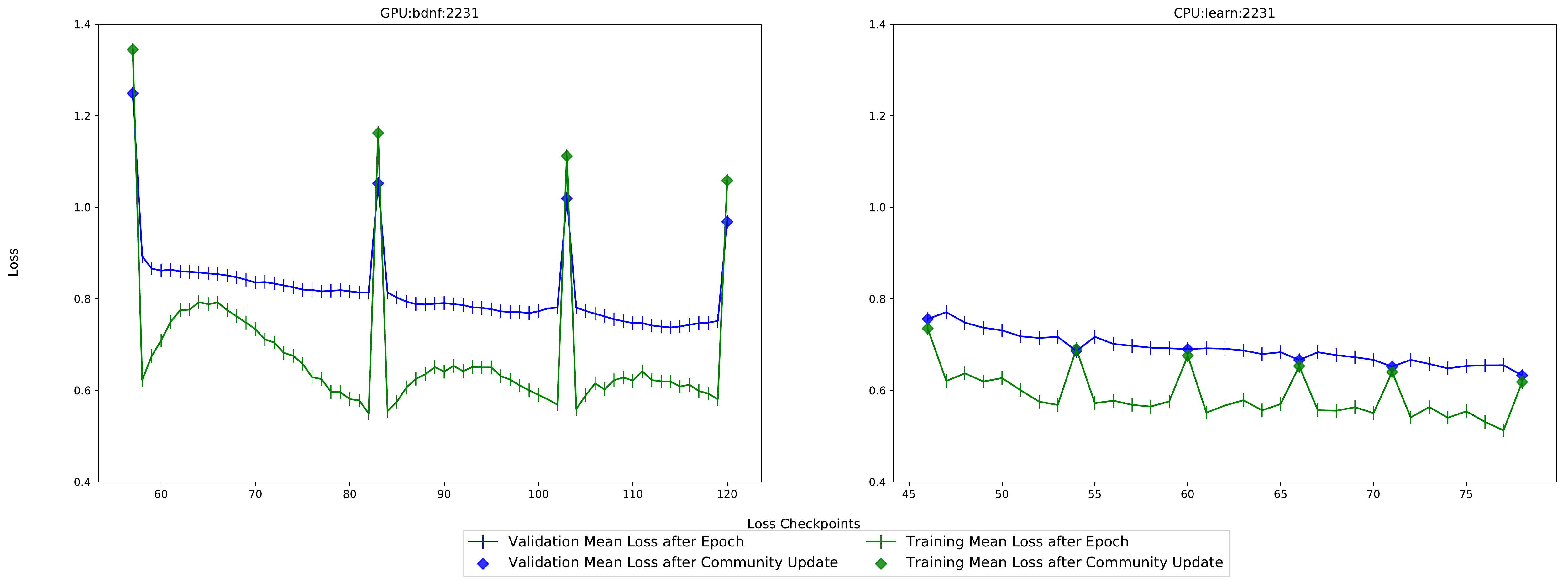}
    \captionsetup{justification=centering}
    \caption{Sample Validation Cycles for a fast (left) and a slow (right) learner, with
    $VC_{Loss}:0,1$ (0 for fast, 1 for slow) and $VC_{Tomb}:4,1$ (4 for fast, 1 for slow).}
    \label{fig:ValidationCyclesExample}
\end{figure}

To further control the length of each validation cycle, we introduce an additional sentinel variable, called \textit{Tombstones} ($VC_{Tomb}$), which allows a learner to continue training within a validation cycle even when one of the above two conditions is met. Essentially, the Tombstone threshold is a non negative user-defined hyperparameter which accounts for the total number of failures (i.e., when one of the two conditions is satisfied) that are allowed to occur during a validation cycle. Figure~\ref{fig:ValidationCyclesExample} depicts the validation cycles of a fast (GPU) and a slow (CPU) learner on Cifar10 for an IID and uniform data distribution with the values of the two thresholds for the two computational groups, fast (first value) and slow (second value), being equal to $VC_{Loss}\text{: 0,1}$ and $VC_{Tomb}\text{: 4,1}$. 




\textbf{Staleness Criterion.}  In Asynchronous Federated Learning environments, where a non strict consistency model \cite{lamport1979make} exists, it is inevitable for operations to perform on \textit{stale} models. Community model updates are not directly visible to all learners and different staleness degrees may be observed \cite{ho2013more,cui2014exploiting}. Staleness can be controlled by tracking the total number of iterations or number of steps (i.e., mini-batches) applied on the community model \cite{dai2018toward,dai2018learning}. 

We extend the \textit{effective staleness} definition in \cite{dai2018learning} as follows. For a learner $k$ requesting a community update $u$ at timestamp $t^{\prime}$, its effective staleness $\mathcal{S}_k$, is equal to the total number of steps used in \textit{committed} updates between $t$ and $t^\prime$, including those in update $u$, that is, $\mathcal{S}_k=\mathtt{S}_c^{t^{\prime}} - \mathtt{S}_c^{t} + \mathtt{S}_k$, with $\mathtt{S}_c^{t^{\prime}}$ denoting the total number of committed steps to the latest community model, $\mathtt{S}_c^{t}$ the total number of committed steps to the community model at timestamp $t$, and $\mathtt{S}_k$ the total number of steps performed by learner $k$ on its local model between timestamps $t$ and $t^\prime$.


\begin{figure}
    \includegraphics[width=\linewidth]{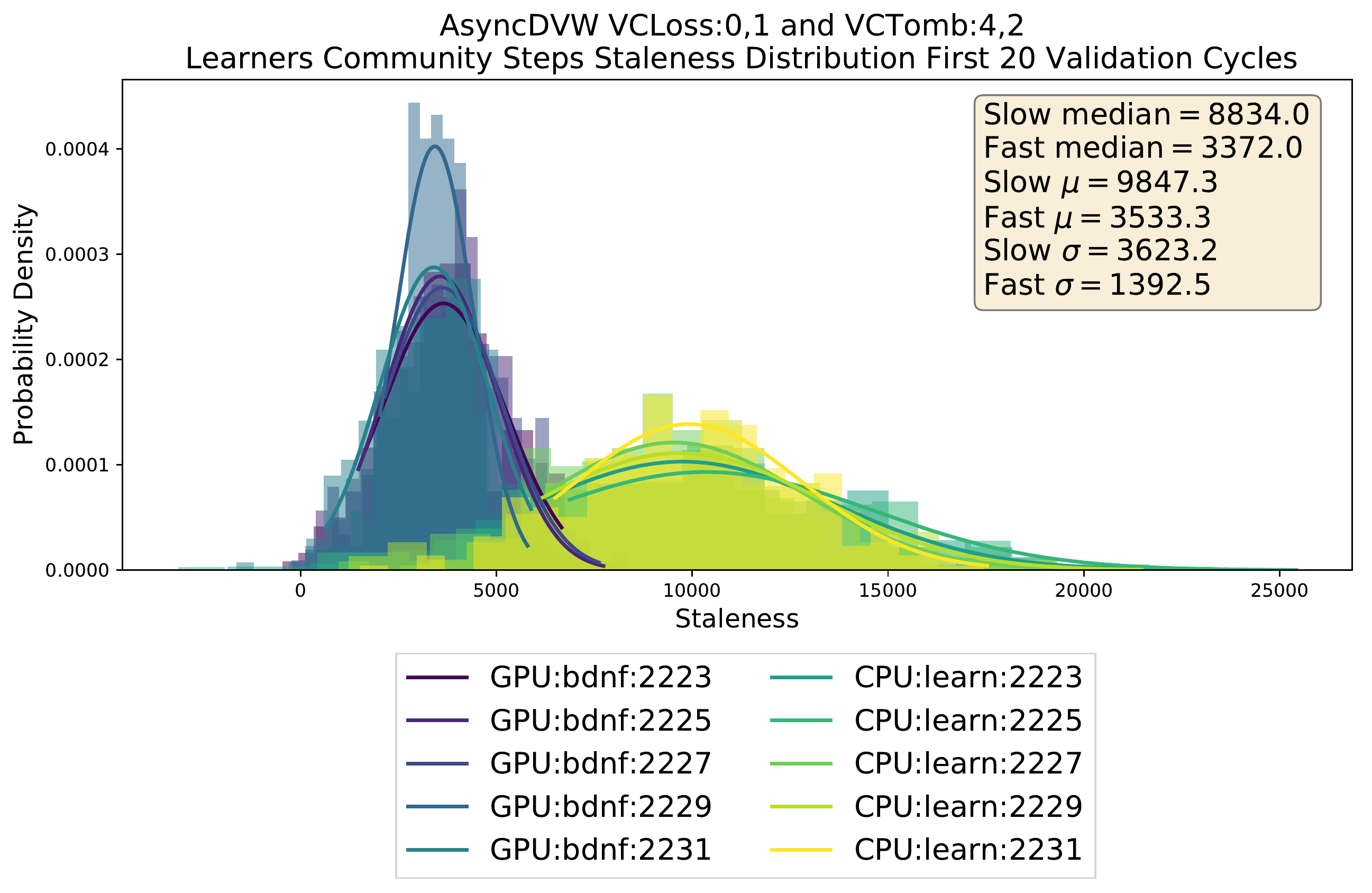}
\captionsetup{justification=centering}
\caption{Staleness distribution for fast and slow learners.}
\vspace*{-5mm}
\label{fig:StalenessDistribution}
\end{figure}

For each learner and every validation cycle we record its effective staleness between community update requests. Once the learner completes its first 20 validation cycles, we generate its staleness frequency histogram using the collected values. With the learned distribution we can control staleness on a per learner basis and trigger a community update when it exceeds some threshold value. In our setting, this threshold is the median of the staleness distribution, which leads to the stopping condition:

\textbf{(C3)} $\mathcal{S}_k > Median(\{\mathcal{S}_1^{vc}, \mathcal{S}_2^{vc}, \ldots, \mathcal{S}_n^{vc} \mid \mathcal{S}_i^{vc}\geq0 \})$

\noindent with $\mathcal{S}_k$ being the current effective staleness of learner $k$ and $\mathcal{S}_i^{vc}$ the effective staleness of validation cycle $i$. Figure \ref{fig:StalenessDistribution} shows the staleness distribution of the first 20 validation cycles for each learner under an IID and uniform data distribution, including the median, mean and standard deviation of staleness for each computational group. As shown, learners that belong to the same computational group (GPU and CPU) display similar levels of staleness.

\section{Evaluation}\label{sec:Evaluation}

We provide an experimental analysis of our proposed DVW weighting scheme under synchronous and asynchronous communication protocols (SyncDVW and AsyncDVW), reserving a stratified 5\% of each local training dataset for the creation of the distributed validation dataset and using Micro F\textsubscript{1}-Score (Sect.~\ref{sec:DVWScheme}) as our assessment method, versus both synchronous and asynchronous FedAvg (SyncFedAvg and AsyncFedAvg) and FedAsync \cite{xie2019asynchronous}, which use 100\% of the training data. 
FedAsync was run using the polynomial staleness function, i.e., FedAsync$+$Poly, with mixing hyperparameter $a=0.5$ and model divergence regularization factor $\rho=0.005$, which is reported to exhibit the best performance in \cite{xie2019asynchronous}.  
AsyncDVW and AsyncFedAvg used the caching mechanism described in Section~\ref{sec:MetisFederatedLearningFramework} (but not FedAsync). The weighting scheme of AsyncFedAvg is exactly the same as SyncFedAvg, i.e., the learners local model contribution in the community is based on the size of their training set.
We evaluate the federation policies on the Cifar10, Cifar100 and ExtendedMNIST By Class \cite{cohen2017emnist} domains. The first two are representative cases of balanced domains, namely an equal number of data examples per class exist in training and test sets, while the latter is a representative case of imbalanced domains, in which a different number of data examples per class exist in training and test.  

\textbf{Models Architecture.} The architecture of the deep learning networks 
for Cifar10 and Cifar100
come from the Tensorflow tutorials, and for ExtendedMNIST come from the LEAF benchmark \cite{caldas2018leaf}. For Cifar10 and ExtendedMNIST we train a 2-CNN, while for Cifar100 we train a ResNet-50.\footnote{Cifar10:\url{https://github.com/tensorflow/models/tree/master/tutorials/image/cifar10}, Cifar100:\url{https://github.com/tensorflow/models/tree/r1.13.0/official/resnet}, EMNIST:\url{https://github.com/TalwalkarLab/leaf/blob/master/models/femnist/cnn.py}}
For all models, during training, we share all trainable weights (i.e., kernels and biases). For ResNet we also share the batch normalization, gamma and beta matrices. The random seed for all our experiments is set to 1990.


\textbf{Models Hyperparameters.} For all the neural networks, we initially performed a grid search over the \textit{centralized} model to identify the hyperparameters with the best performance. The optimization method used for training all the models is Stochastic Gradient Descent (SGD) with Momentum. We performed a grid search over a range of learning rate, $\eta$, momentum factor, $\gamma$, and mini batch size, $\beta$, values. After identifying the optimal combination, we kept the hyperparameter values fixed throughout the federation training. In particular, we did not apply any learning rate annealing schedule (but see Section~\ref{sec:Discussion}). Explicitly, for Cifar10 we used, $\eta$=0.05, $\gamma$=0.75 and $\beta$=100, for Cifar100, $\eta$=0.1, $\gamma$=0.9, $\beta$=128 and for ExtendedMNIST, $\eta$=0.01, $\gamma$=0.5 and $\beta$=100.

\textbf{Learners Update Frequency.} For both synchronous and asynchronous policies, we evaluated the convergence rate of the federation under different update frequencies (i.e., number of local epochs before triggering a community update request). For synchronous and asynchronous FedAvg, and for synchronous DVW, we evaluated several update frequencies $uf=\{1,2,4,8,16,32\}$. We use $uf=4$ for those policies, since it had the best performance. We refer to DVW with static update frequency as DVW non-adaptive[na].
For asynchronous DVW, we performed a grid search over the validation cycles, validation loss $VC_{Loss}$, and tombstones $VC_{Tomb}$ thresholds. The staleness criterion is determined adaptively once a learner finishes its first 20 validation cycles. During the execution of the asynchronous DVW policy, a learner makes a community update request when one of the two stopping criteria, Validation Loss or Staleness, is reached; whichever is satisfied first (cf. Section \ref{sec:StoppingCriteriaAsyncDVW}). Empirically, the Validation Loss criterion triggers the majority of the update requests. We refer to DVW with dynamic update frequency as DVW adaptive[a].


\textbf{Computational Environment.} Our homogeneous federation environment consists of 10 fast learners (GPU). Our heterogeneous environment consists of 5 fast (GPU) and 5 slow (CPU) learners. The fast learners run on a dedicated GPU server equipped with 8~GeForce GTX 1080 Ti graphics cards of 10~GB RAM each, 40~Intel(R) Xeon(R) CPU E5-2630 v4 @ 2.20GHz, and 128GB DDR4 RAM. The slow learners run on a separate server equipped with 48 Intel(R) Xeon(R) CPU E5-2650 v4 @ 2.20GHz and 128GB DDR4 RAM.

\textbf{Data Distributions.} We evaluate the federation policies over multiple training datasets with heterogeneous data sizes and class distributions.\footnote{The distributions of our experiments are at: \url{https://dataverse.harvard.edu/privateurl.xhtml?token=90a0245c-32c4-42bf-8edc-c83d9217deee}}
We consider three types of data size distributions: \textit{Uniform}, where every learner has the same number of examples; \textit{Skewed}, where the distribution of the number of examples is rightly skewed and hence no learner has the exact same data size as the others; and \textit{Power Law}, with the power law's exponent set to~1.5. For class distribution, we assign a different number of examples per class per learner for each domain independently. Specifically, with \textit{IID} we denote the case where all learners hold training examples from all the target classes, and with \textit{Non-IID(x)} we denote the case where every learner holds training examples from only x classes. 
For example, Non-IID(3) in Cifar10 means that each learner only has training examples from 3 target classes (out of the 10 classes in Cifar10). 

For power law data sizes and Non-IID configurations, in order to preserve scale invariance, we needed to assign data from more classes to the learners at the head of the distribution. For example, for Cifar10 with power law and a goal of 5 classes per learner, the actual distribution is Non-IID(8x1,7x1,6x1,5x7), meaning that the first learner holds data from 8 classes, the second from 7 classes, the third from 6 classes, and all 7 subsequent learners hold data from 5 classes. For brevity, we refer to this distribution as Non-IID(5x7).
Similarly for Cifar10 power law and Non-IID(3), the actual distribution is Non-IID(8x1,4x1,3x8), abbreviated to Non-IID(3x8). For Cifar100, power law and Non-IID(50), the actual distribution is Non-IID(84x1,76x1,68x1,64x1,55x1,50x5), abbreviated to Non-IID(50x5).

\begin{figure*}[htpb]
  \centering
  \subfloat[Uniform \& IID]{
    \centering\includegraphics[width=0.32\linewidth]{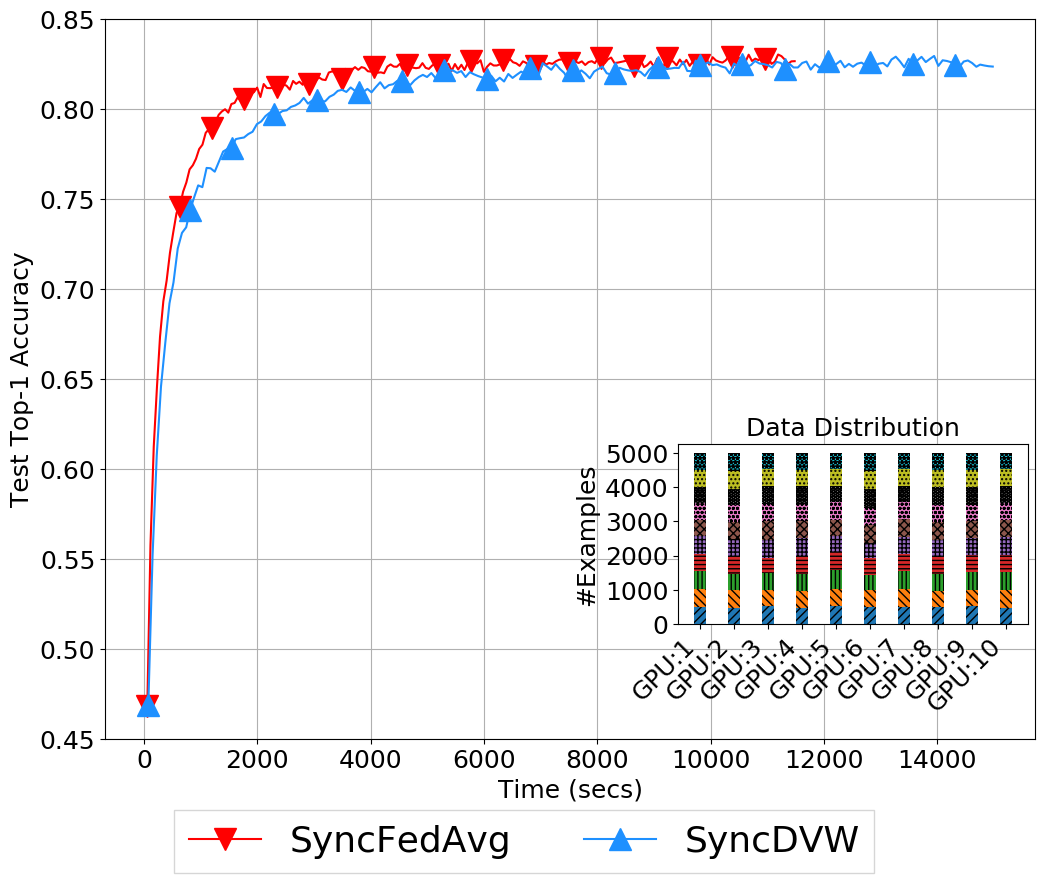}
    \label{subfig:Cifar10_HomogeneousCluster_Uniform_IID}
  }
  \subfloat[Uniform \& Non-IID(5)]{
    \centering\includegraphics[width=0.32\linewidth]{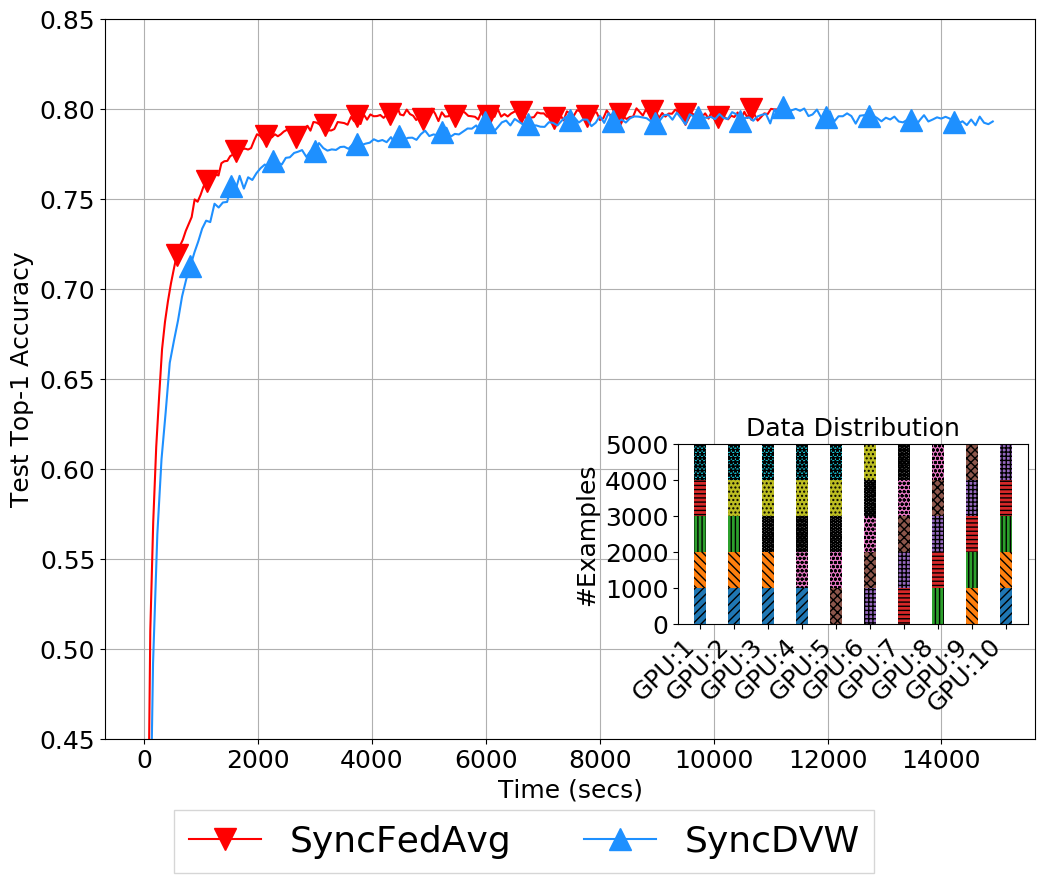}
    \label{subfig:Cifar10_HomogeneousCluster_Uniform_NonIID_5}
  }
  \subfloat[Uniform \& Non-IID(3)]{
    \centering\includegraphics[width=0.32\linewidth]{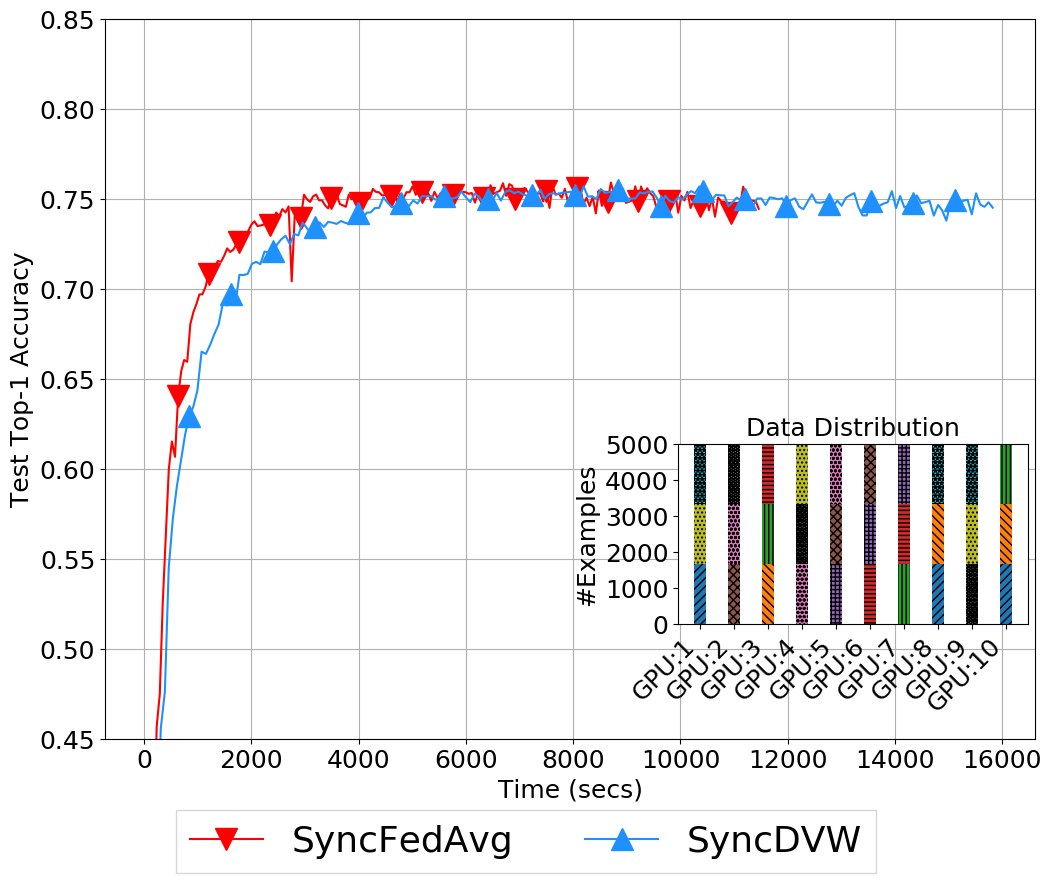}
    \label{subfig:Cifar10_HomogeneousCluster_Uniform_NonIID_3}
  }
  
  \subfloat[Skewed \& IID]{
    \centering\includegraphics[width=0.32\linewidth]{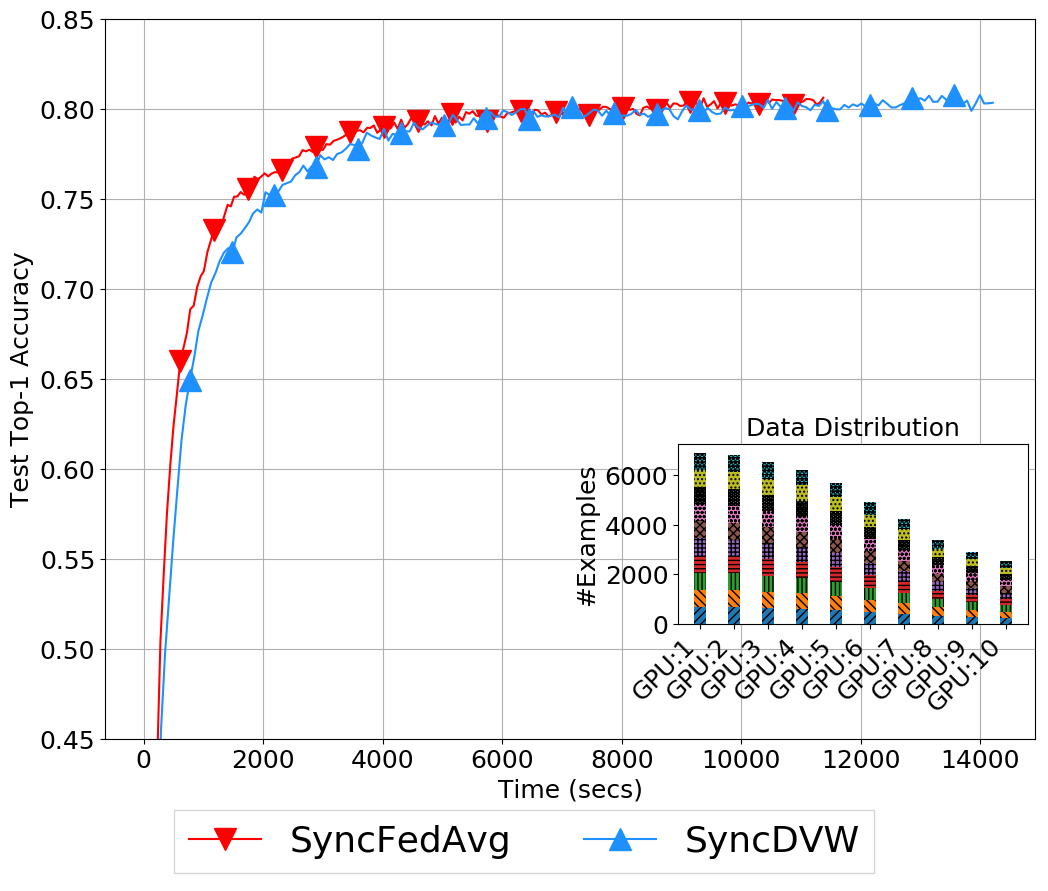}
    \label{subfig:Cifar10_HomogeneousCluster_Skewed_IID}
  }
  \subfloat[Skewed \& Non-IID(5)]{
    \centering\includegraphics[width=0.32\linewidth]{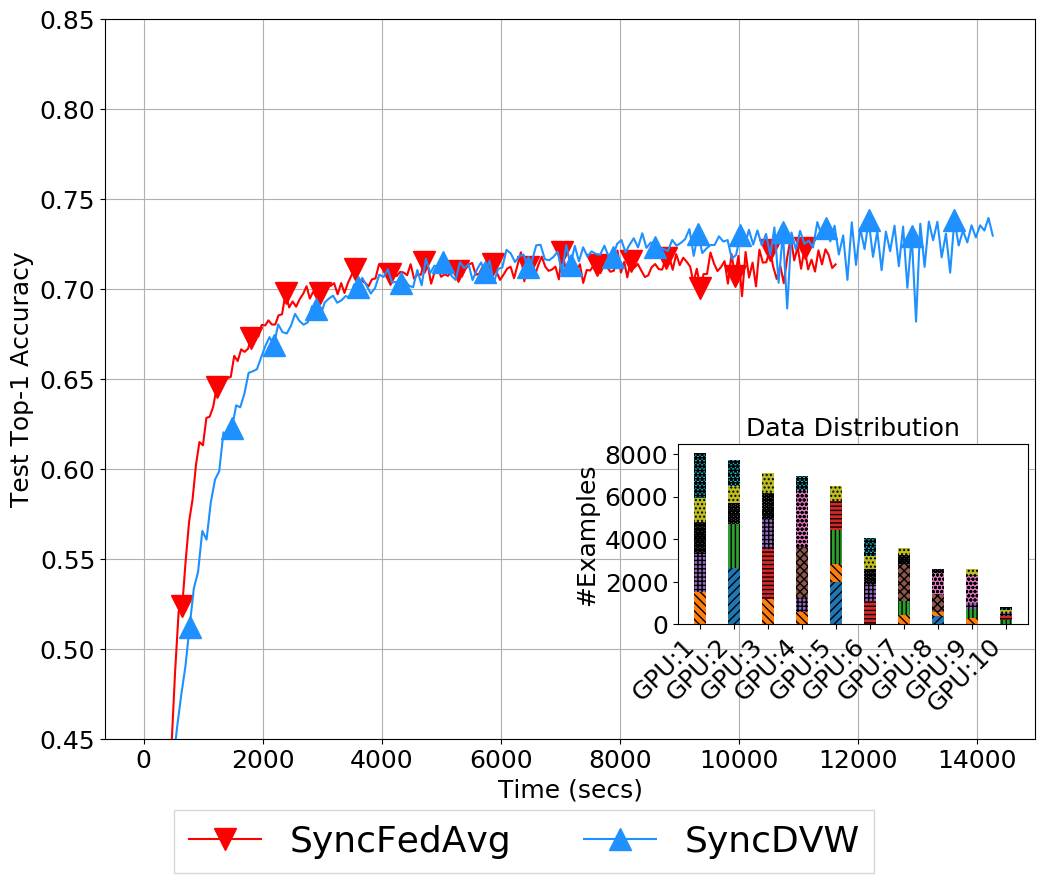}
    \label{subfig:Cifar10_HomogeneousCluster_Skewed_NonIID_5}
  }
  \subfloat[Skewed \& Non-IID(3)]{
    \centering\includegraphics[width=0.32\linewidth]{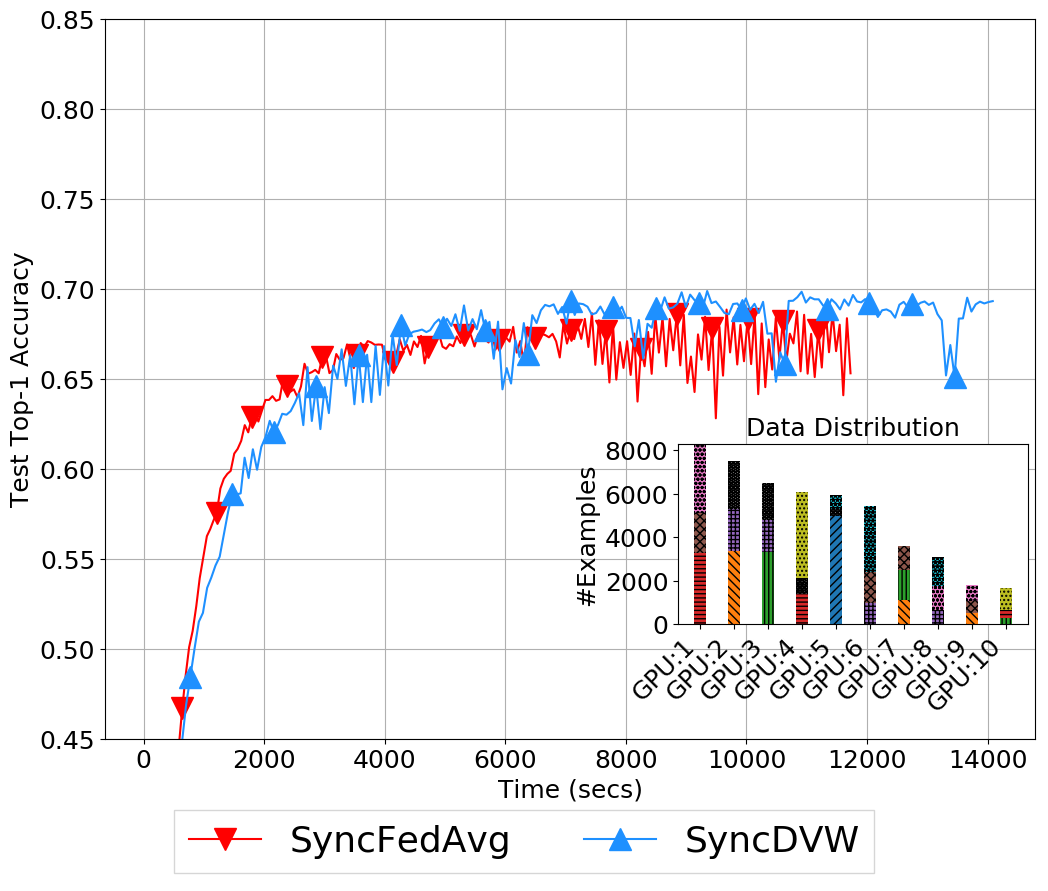}
    \label{subfig:Cifar10_HomogeneousCluster_Skewed_NonIID_3}
  }
  
  \subfloat[Power Law \& IID]{
    \centering\includegraphics[width=0.32\linewidth]{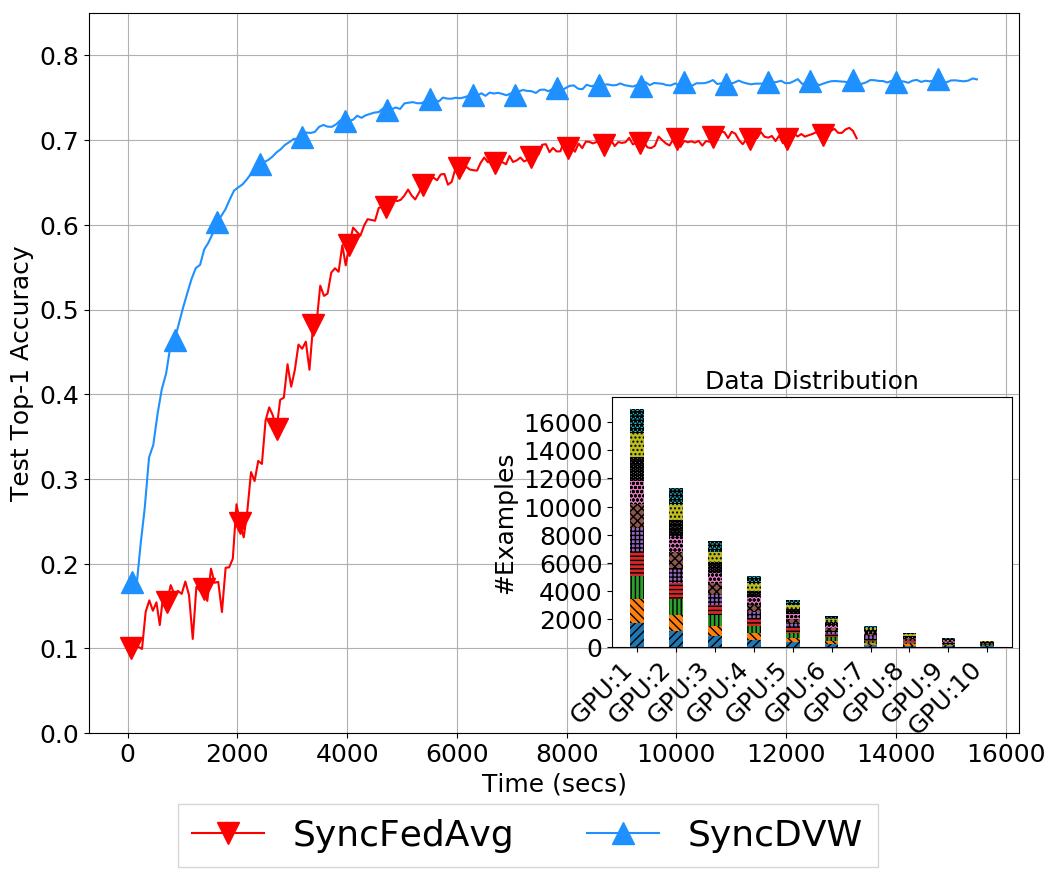}
    \label{subfig:Cifar10_HomogeneousCluster_PowerLaw_IID}
  }
  \subfloat[Power Law \& Non-IID(5x7)]{
    \centering\includegraphics[width=0.32\linewidth]{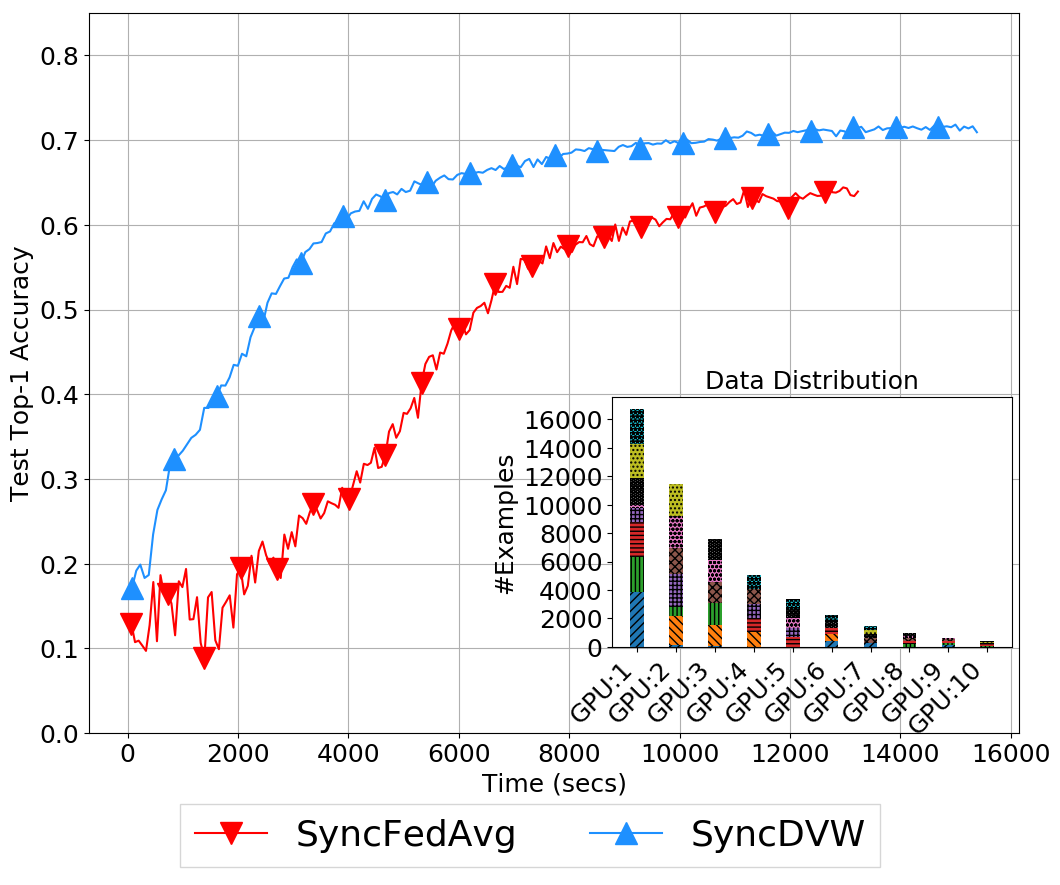}
    \label{subfig:Cifar10_HomogeneousCluster_PowerLaw_NonIID_5}
  }
  \subfloat[Power Law \& Non-IID(3x8)]{
    \centering\includegraphics[width=0.32\linewidth]{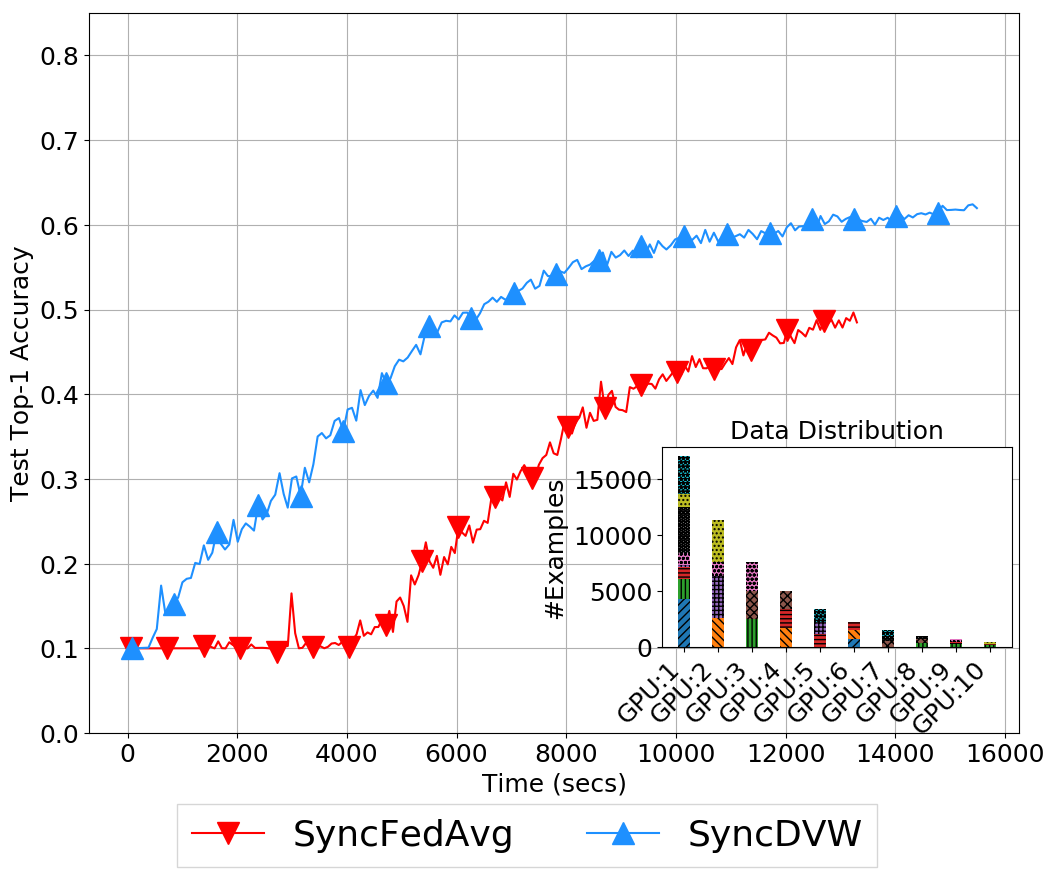}
    \label{subfig:Cifar10_HomogeneousCluster_PowerLaw_NonIID_3}
  }
  
  \captionsetup{justification=centering}
  \caption{Wall-Clock Time Convergence for Cifar10 on a Homogeneous Cluster}
  \label{fig:Cifar10SynchronousHomogeneousCluster}
\end{figure*}

\begin{figure*}[htbp]
  \centering
  \subfloat[Uniform \& IID]{
    \centering\includegraphics[width=0.32\linewidth]{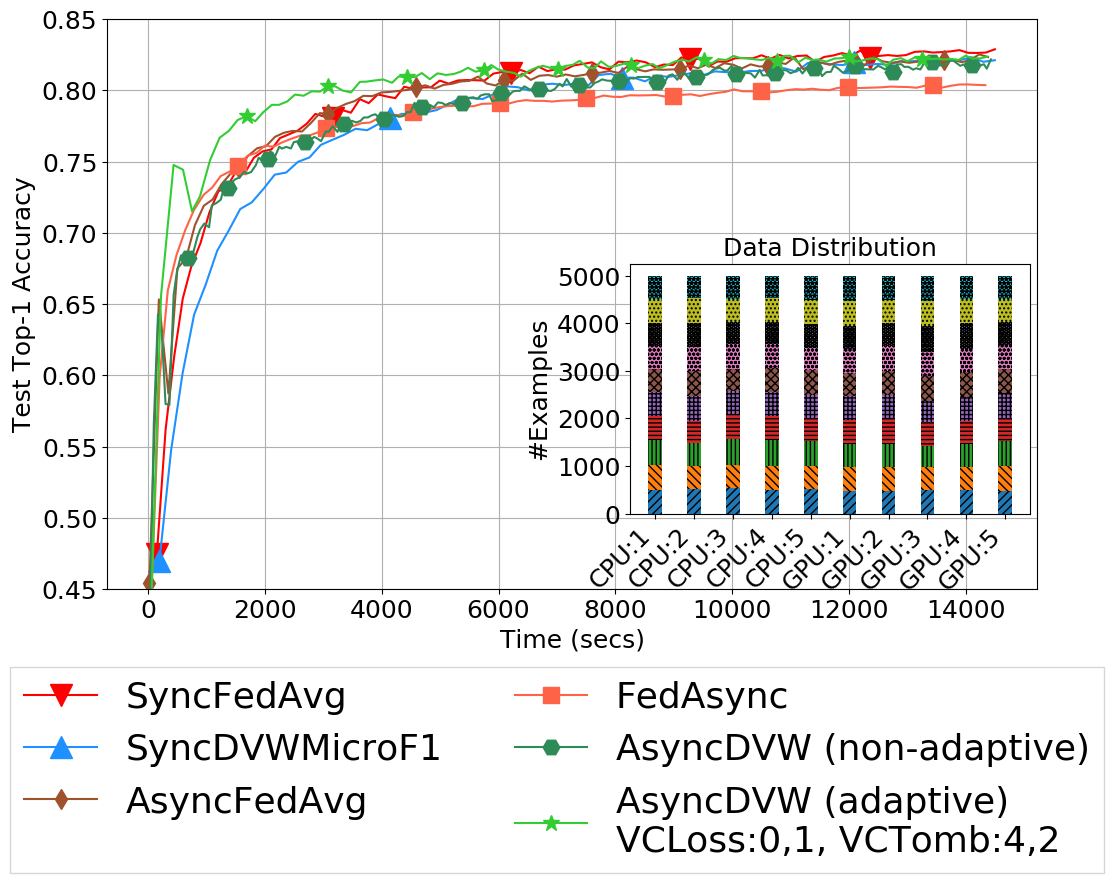}
    \label{subfig:Cifar10_HeterogeneousCluster_Uniform_IID}
  }
  \subfloat[Uniform \& Non-IID(5)]{
    \centering\includegraphics[width=0.32\linewidth]{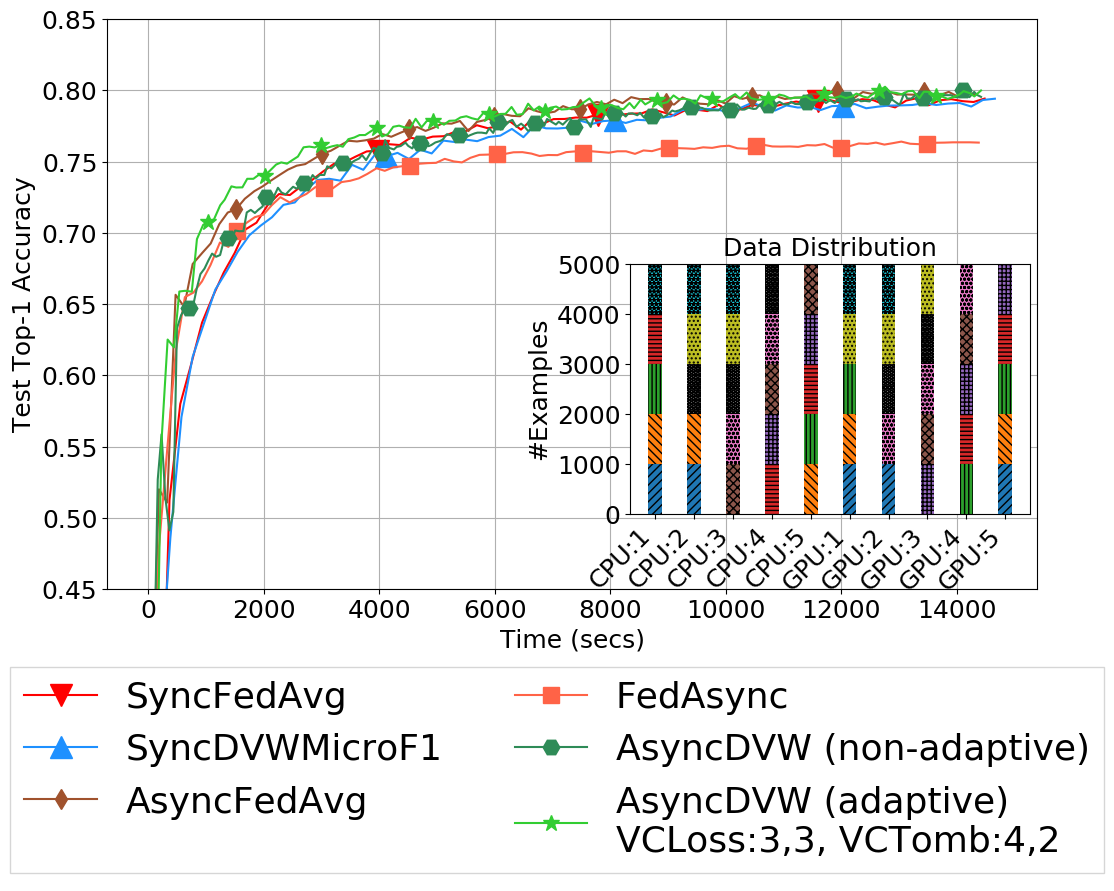}
    \label{subfig:Cifar10_HeterogeneousCluster_Uniform_NonIID_5}
  }
  \subfloat[Uniform \& Non-IID(3)]{
    \centering\includegraphics[width=0.32\linewidth]{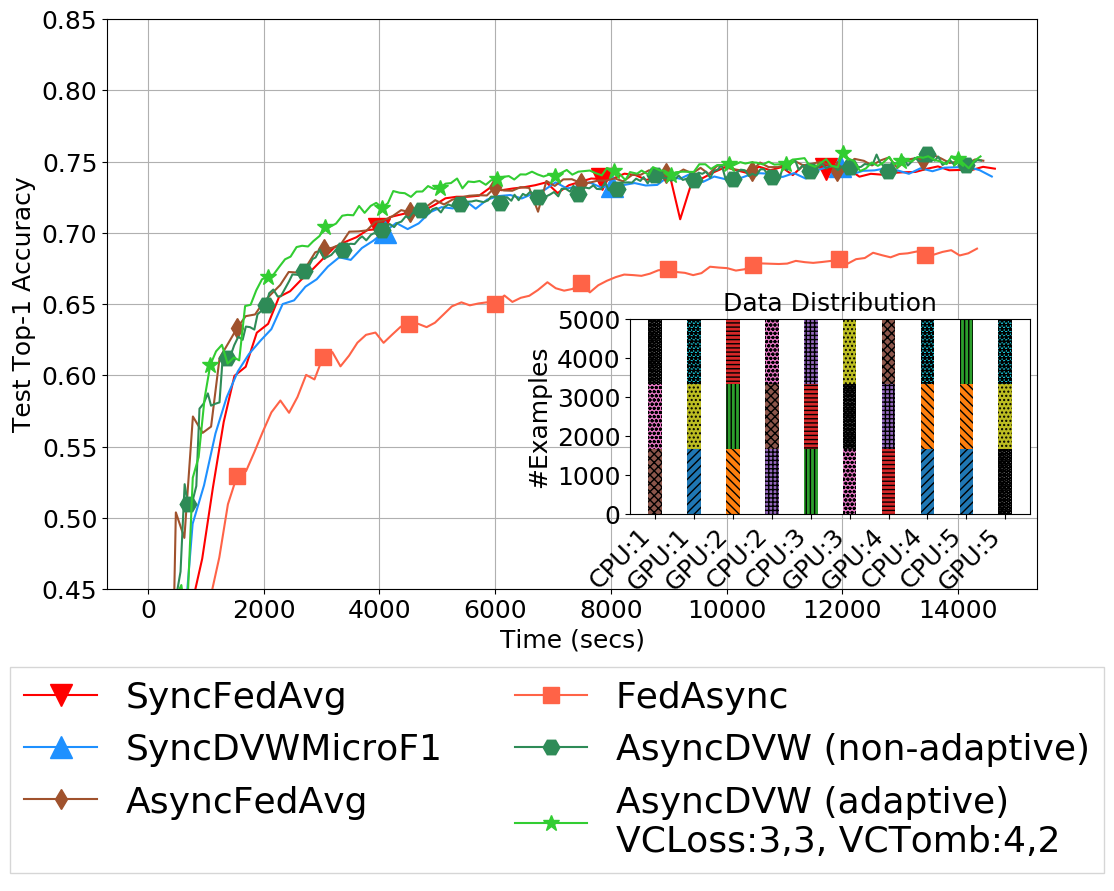}
    \label{subfig:Cifar10_HeterogeneousCluster_Uniform_NonIID_3}
  }
  
  \subfloat[Skewed \& IID]{
    \centering\includegraphics[width=0.32\linewidth]{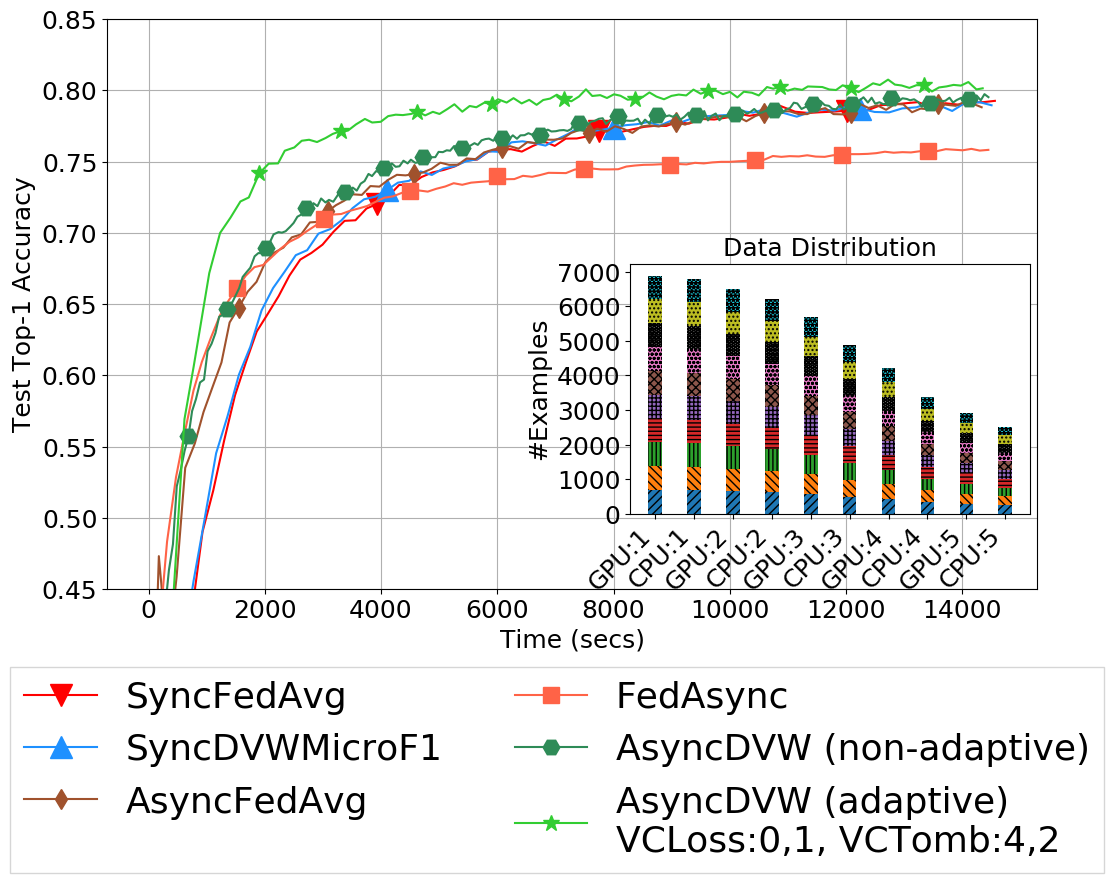}
    \label{subfig:Cifar10_HeterogeneousCluster_Skewed_IID}
  }
  \subfloat[Skewed \& Non-IID(5)]{
    \centering\includegraphics[width=0.32\linewidth]{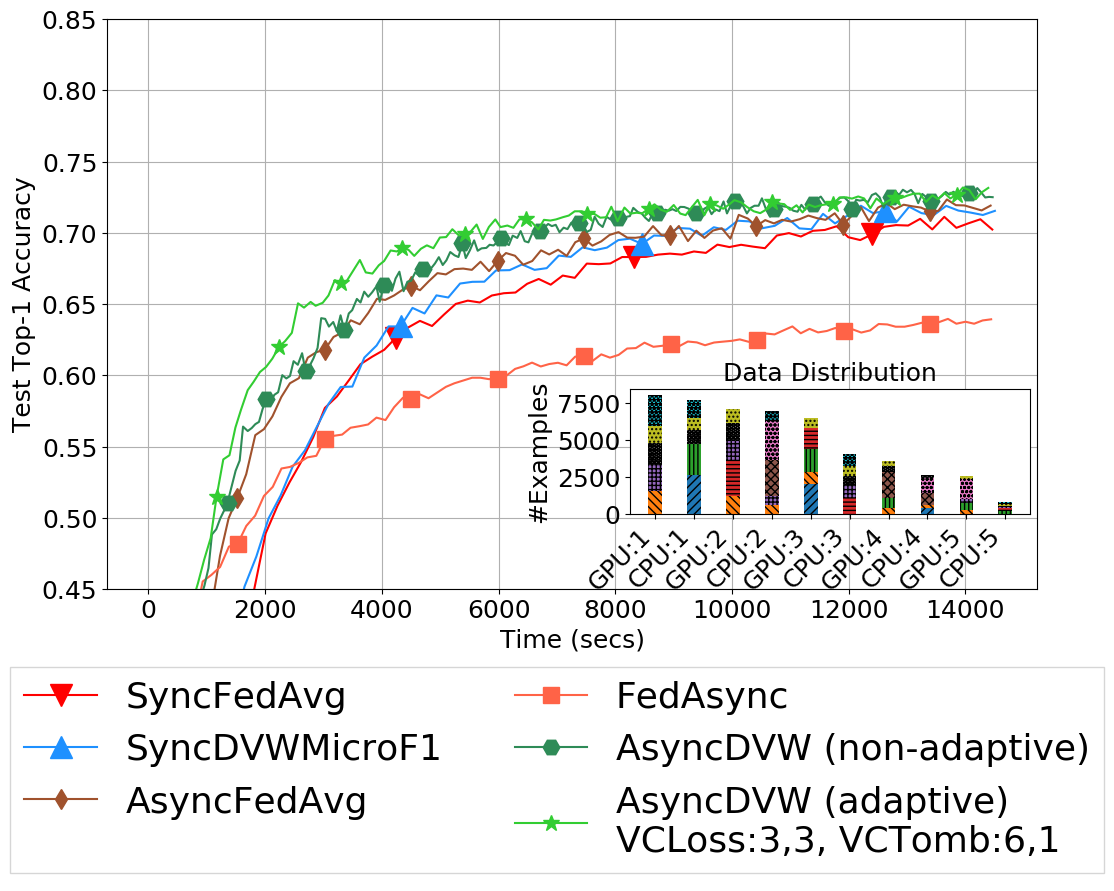}
    \label{subfig:Cifar10_HeterogeneousCluster_Skewed_NonIID_5}
  }
  \subfloat[Skewed \& Non-IID(3)]{
    \centering\includegraphics[width=0.32\linewidth]{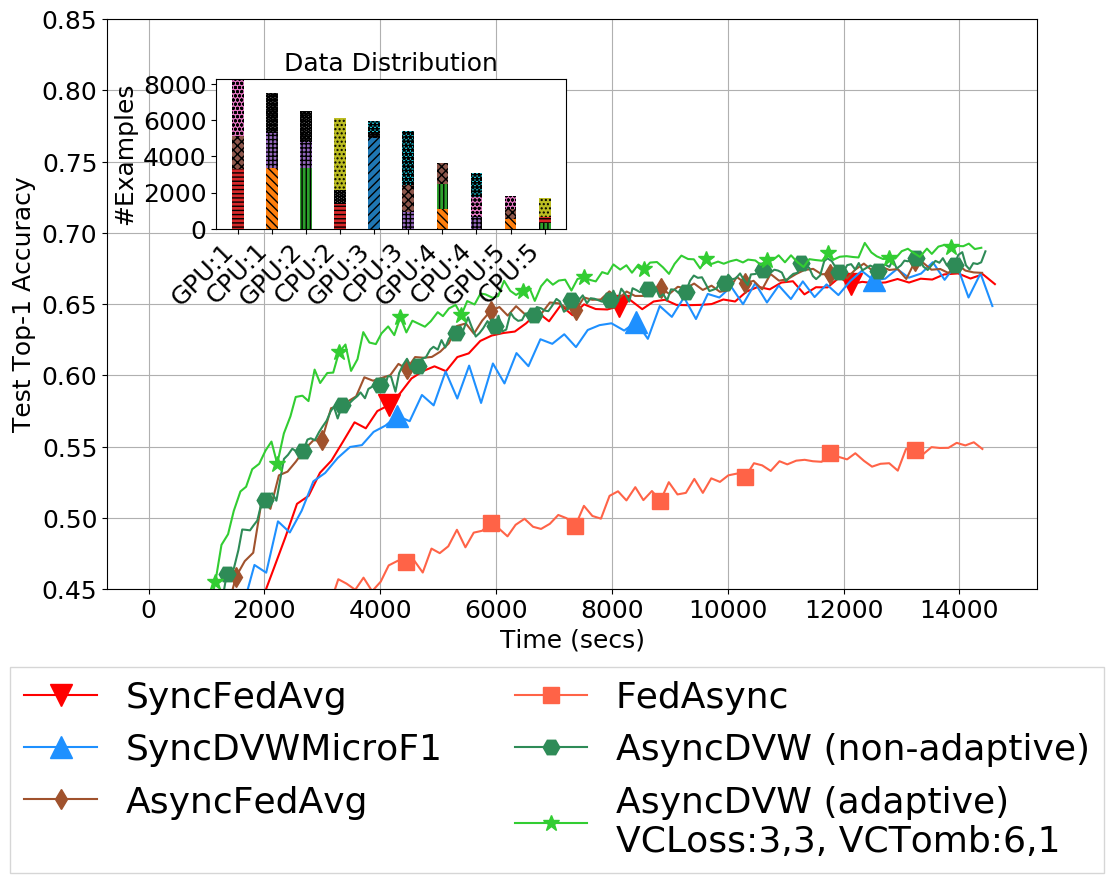}
    \label{subfig:Cifar10_HeterogeneousCluster_Skewed_NonIID_3}
  }
  
  \subfloat[Power Law \& IID]{
    \centering\includegraphics[width=0.32\linewidth]{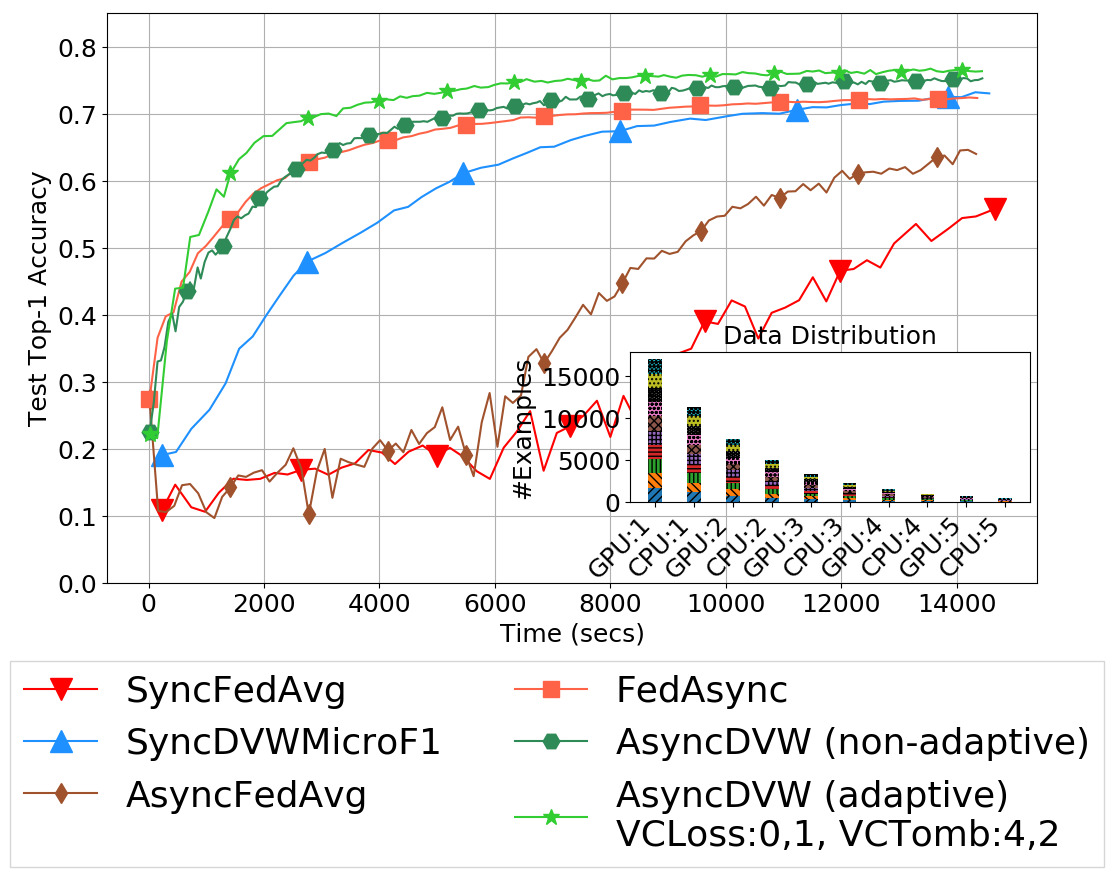}
    \label{subfig:Cifar10_HeterogeneousCluster_PowerLaw_IID}
  }
  \subfloat[Power Law \& Non-IID(5x7)]{
    \centering\includegraphics[width=0.32\linewidth]{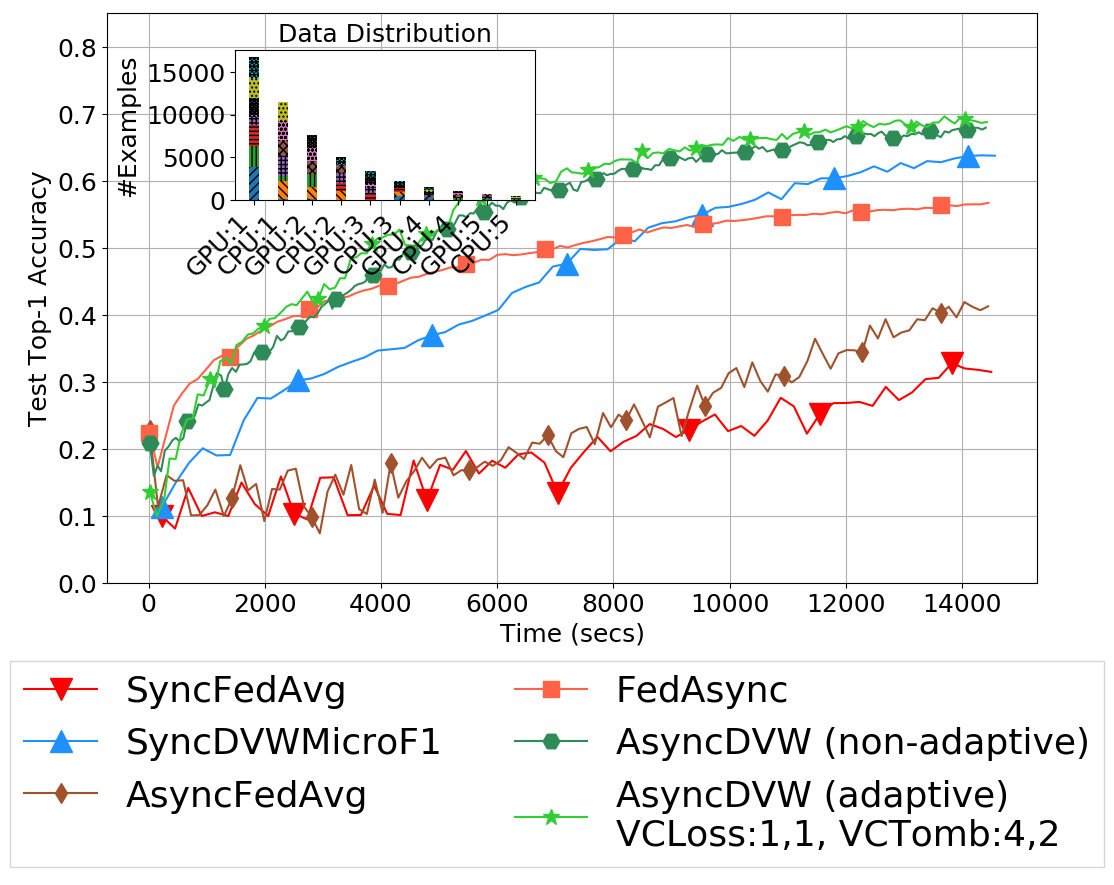}
    \label{subfig:Cifar10_HeterogeneousCluster_PowerLaw_NonIID_5}
  }
  \subfloat[Power Law \& Non-IID(3x8)]{
    \centering\includegraphics[width=0.32\linewidth]{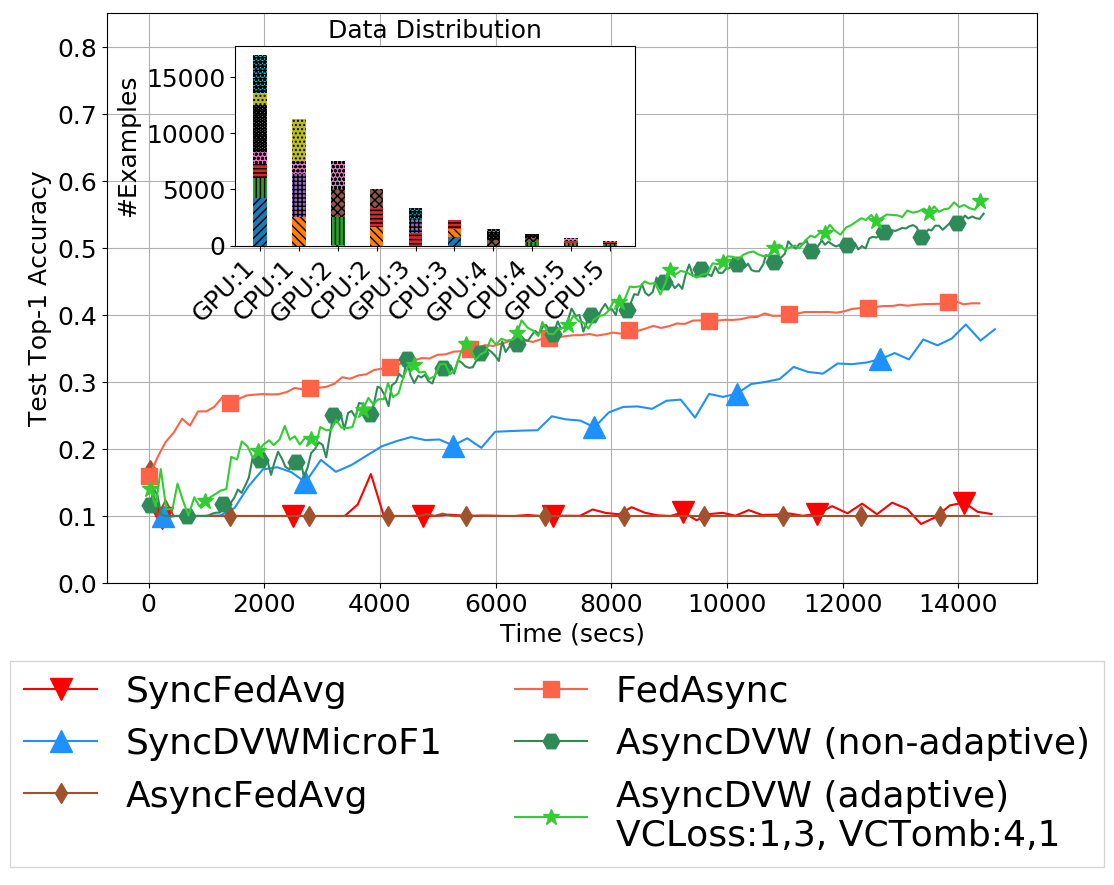}
    \label{subfig:Cifar10_HeterogeneousCluster_PowerLaw_NonIID_3}
  }
  
  \captionsetup{justification=centering}
  \caption{Wall-Clock Time Convergence for Cifar10 on a Heterogeneous Cluster}
  \label{fig:Cifar10SynchronousAsyncrhonousHeterogeneousCluster}
\end{figure*}

\begin{figure*}[htbp]
  \centering
  
  \subfloat[Uniform \& IID]{
    \centering\includegraphics[width=0.32\linewidth]{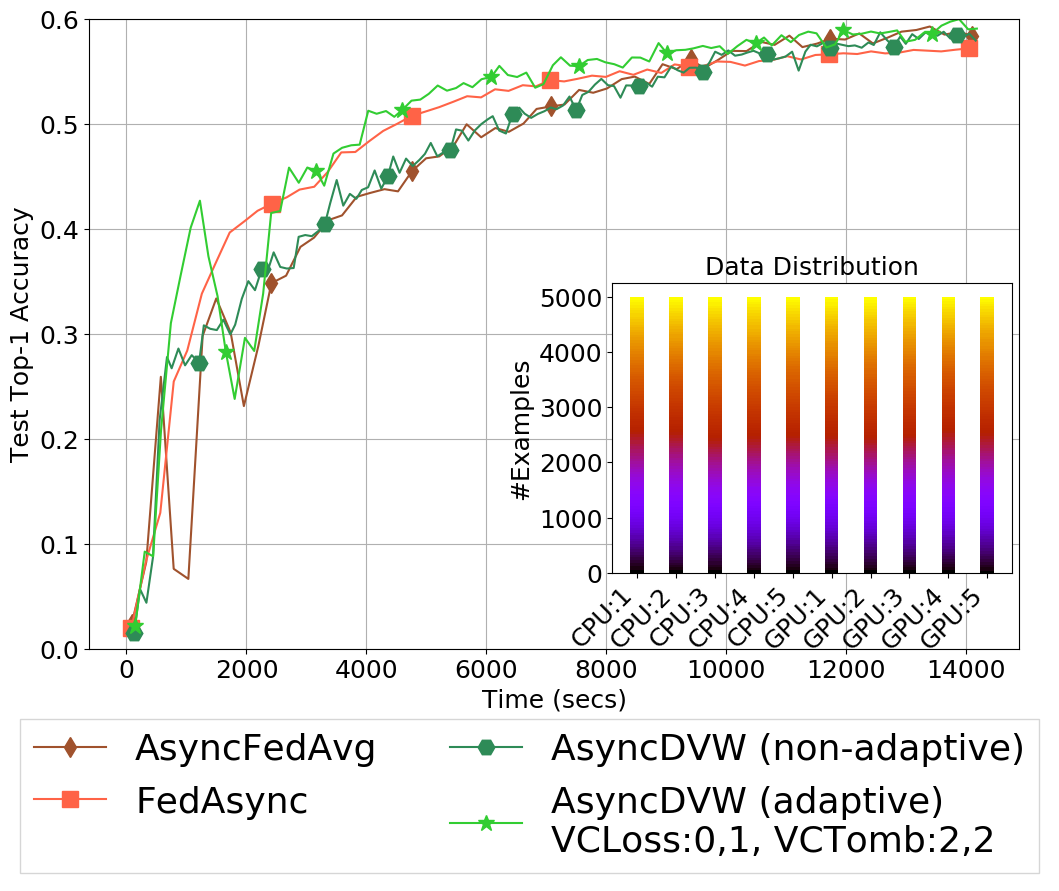}
    \label{subfig:Cifar100_HeterogeneousCluster_Uniform_IID}
  }
  \subfloat[Uniform \& Non-IID(50)]{
    \centering\includegraphics[width=0.32\linewidth]{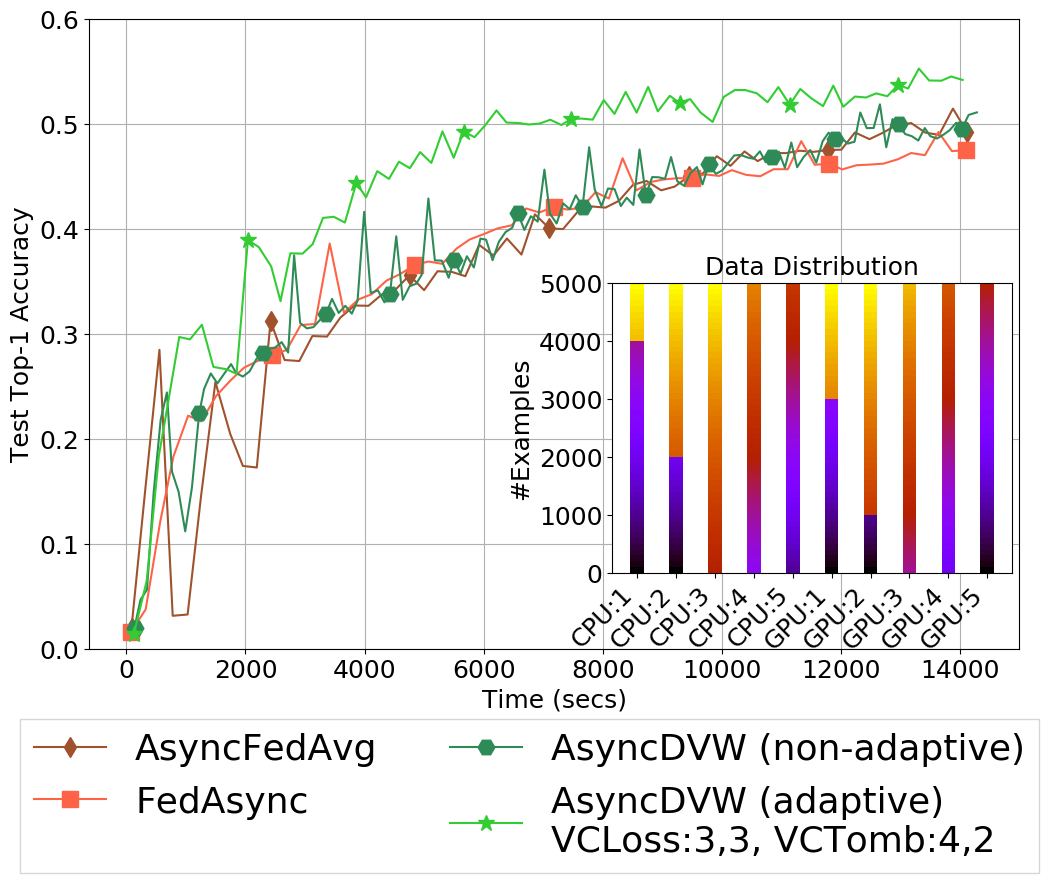}
    \label{subfig:Cifar100_HeterogeneousCluster_Uniform_NonIID_50}
  }
  \subfloat[Skewed \& IID]{
    \centering\includegraphics[width=0.32\linewidth]{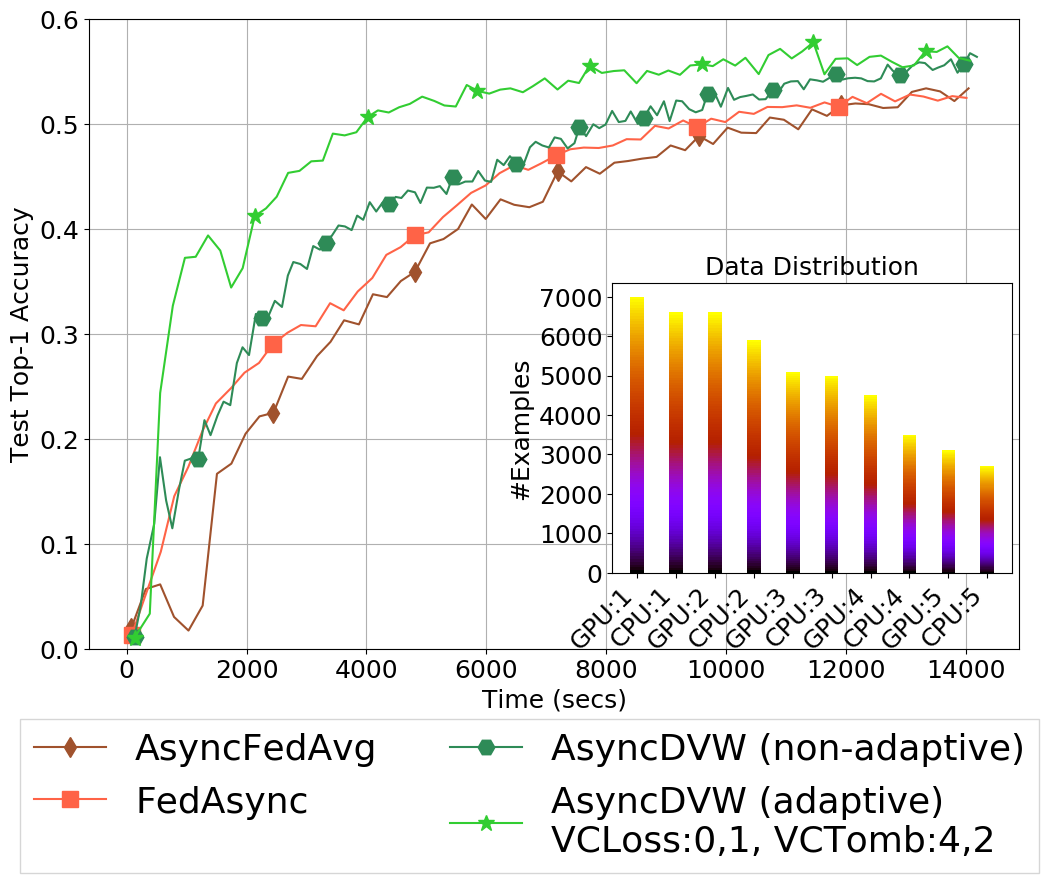}
    \label{subfig:Cifar100_HeterogeneousCluster_Skewed_IID}
  }
  
  \subfloat[Skewed \& Non-IID(50)]{
    \centering\includegraphics[width=0.32\linewidth]{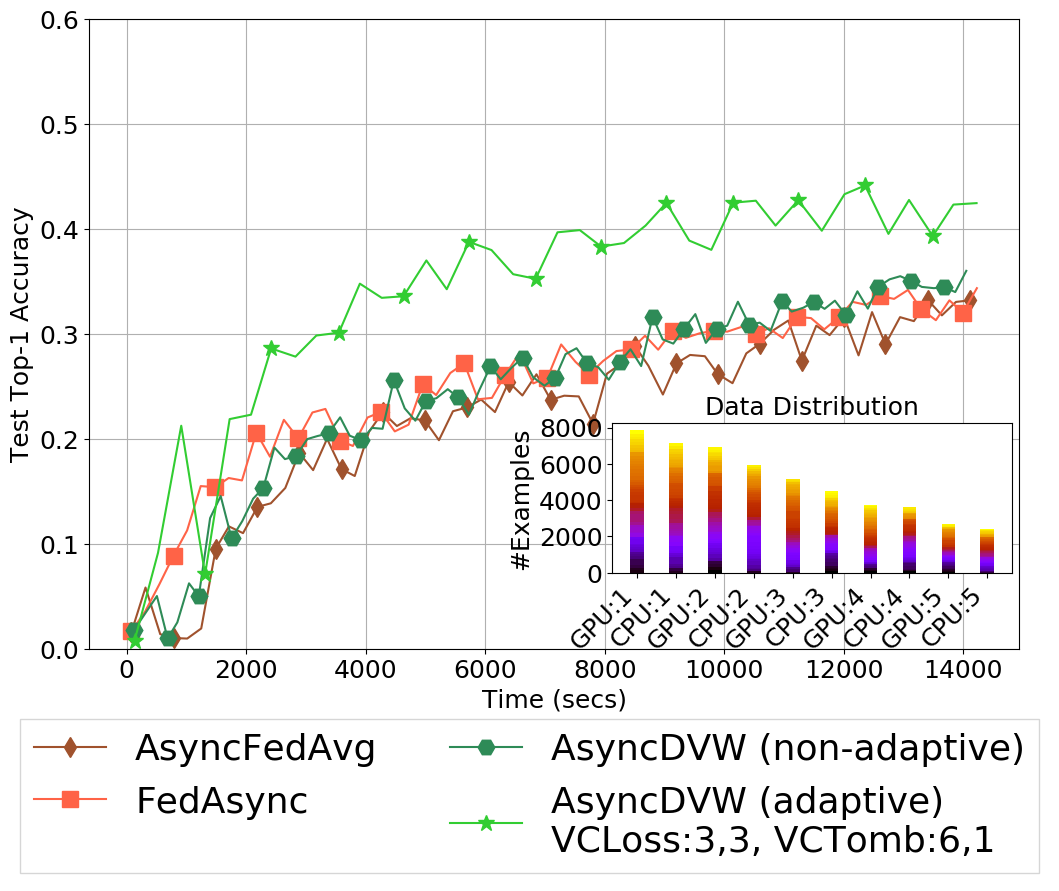}
    \label{subfig:Cifar100_HeterogeneousCluster_Skewed_NonIID_50}
  }
  \subfloat[Power Law \& IID]{
    \centering\includegraphics[width=0.32\linewidth]{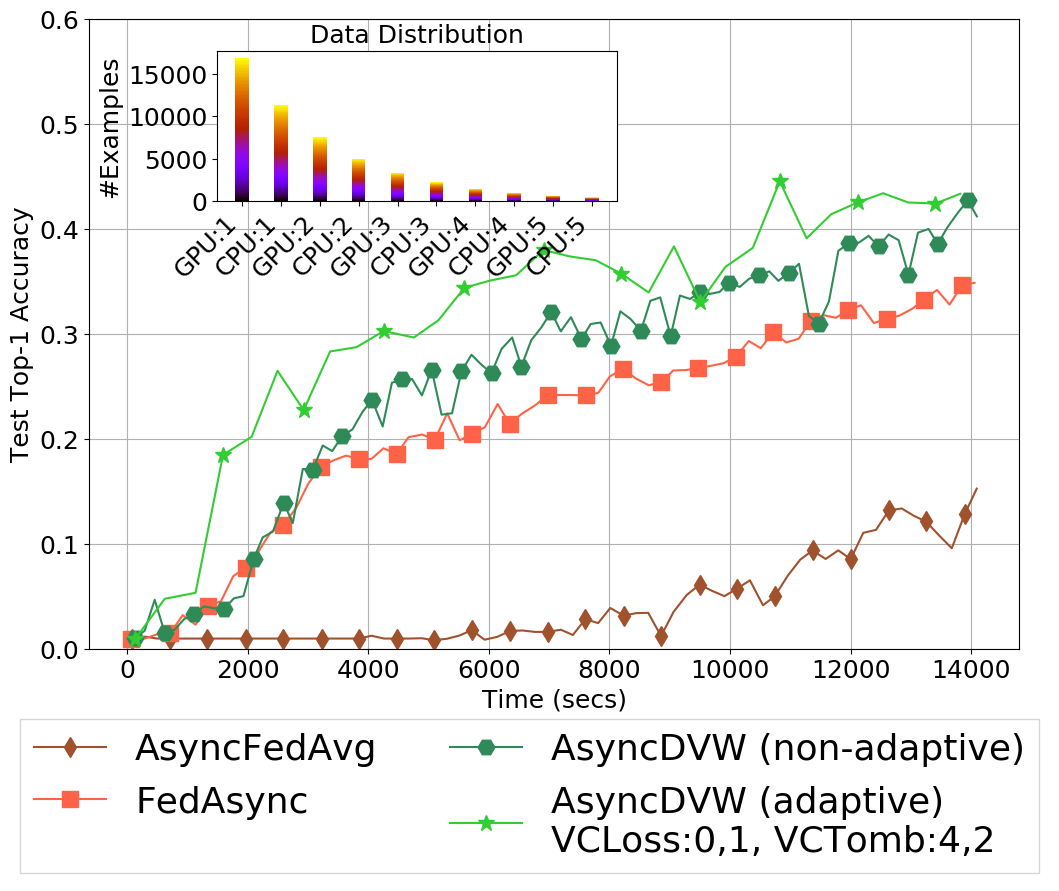}
    \label{subfig:Cifar100_HeterogeneousCluster_PowerLaw_IID}
  }
  \subfloat[Power Law \& Non-IID(50x5)]{
    \centering\includegraphics[width=0.32\linewidth]{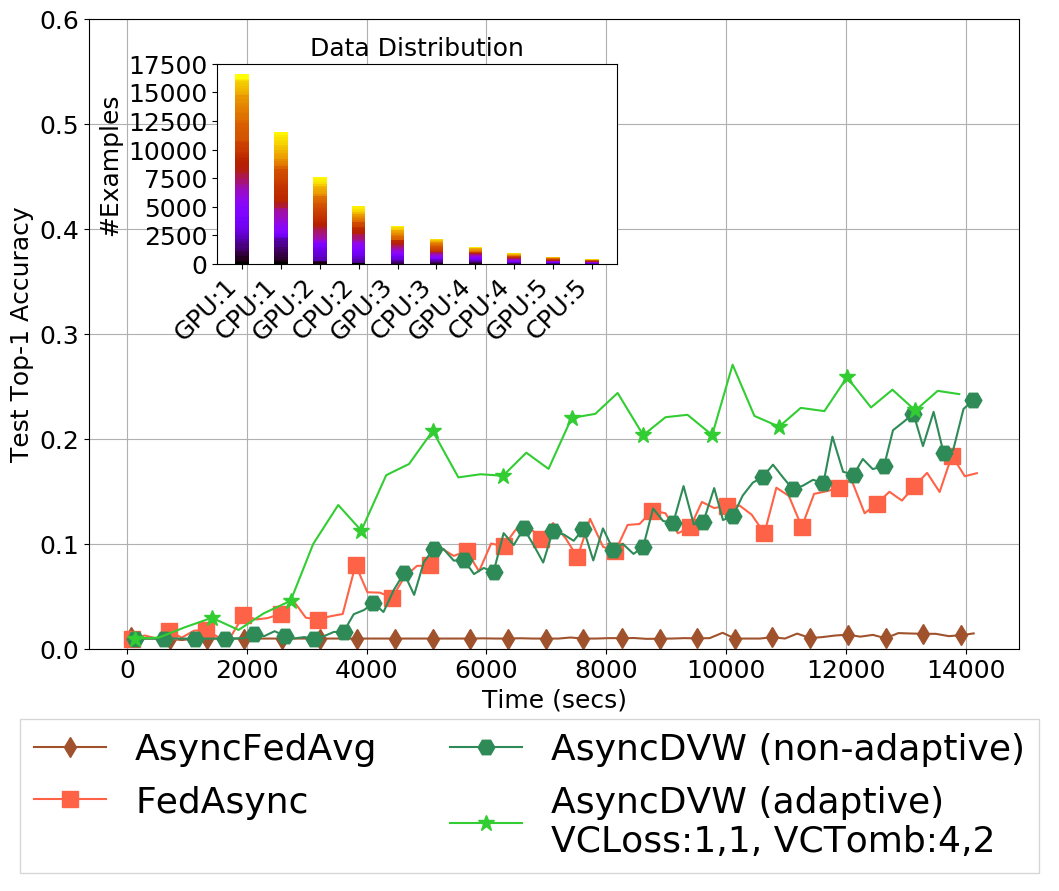}
    \label{subfig:Cifar100_HeterogeneousCluster_PowerLaw_NonIID_50}
  }
  
  \captionsetup{justification=centering}
  \caption{Wall-Clock Time Convergence for Cifar100 on a Heterogeneous Cluster}
  \label{fig:Cifar100AsyncrhonousHeterogeneousCluster}
\end{figure*}

\begin{figure*}[htpb]
  \centering
  \subfloat[Uniform \& IID]{
    \centering\includegraphics[width=0.32\linewidth]{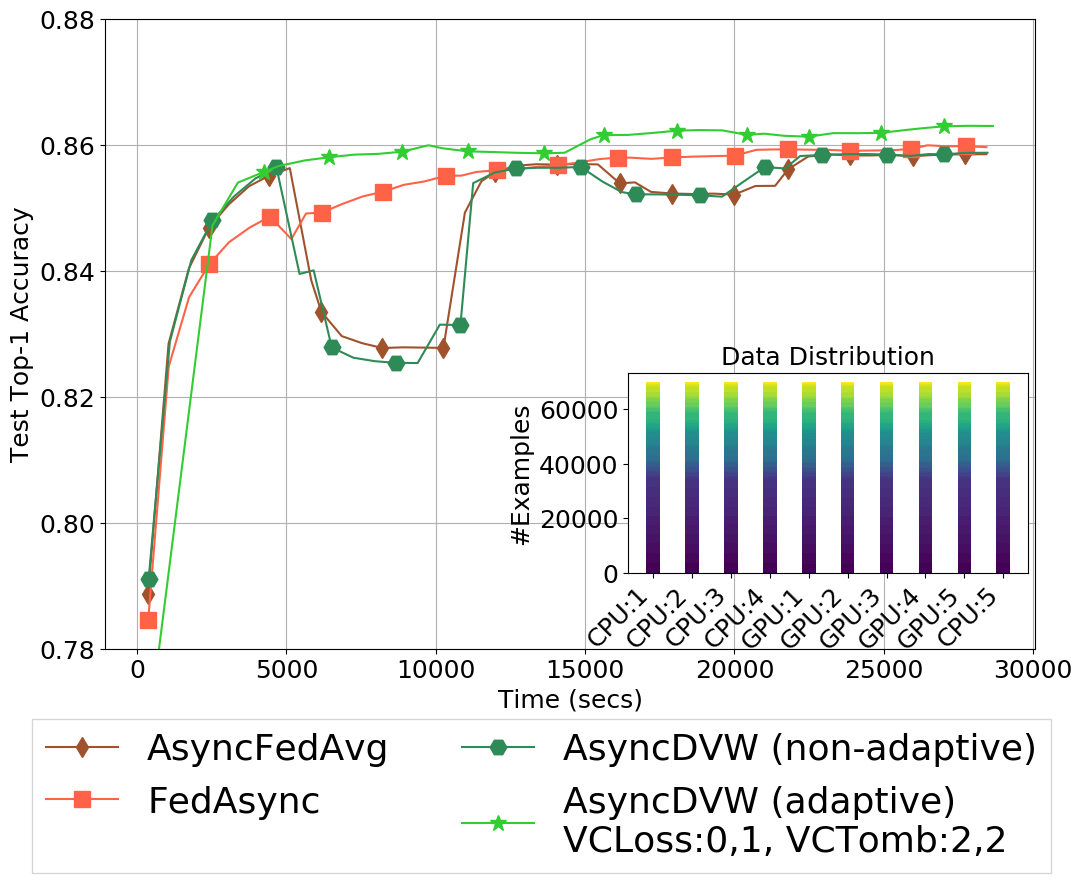}
    \label{subfig:ExtednedMNIST_ByClass_HeterogeneousCluster_Uniform_IID}
  }
  \subfloat[Skewed \& Non-IID(30)]{
    \centering\includegraphics[width=0.32\linewidth]{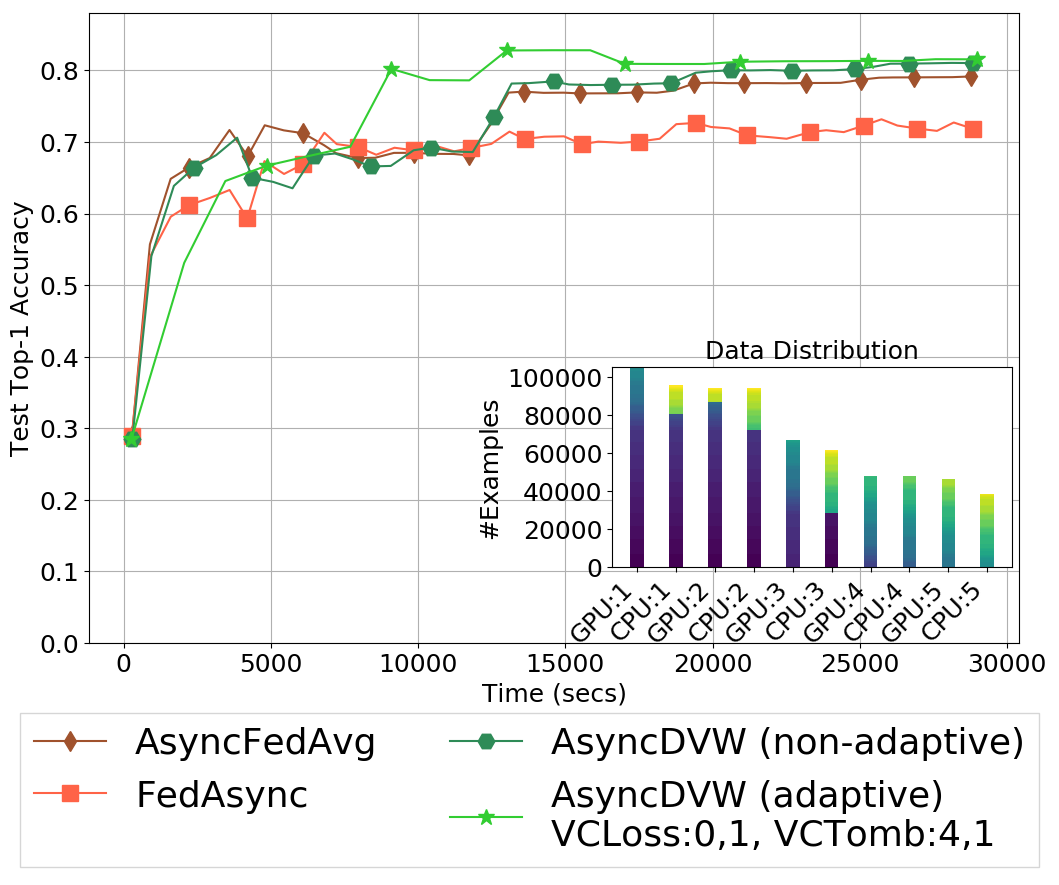}
    \label{subfig:ExtednedMNIST_ByClass_HeterogeneousCluster_Skewed_NonIID_30}
  }
  \subfloat[Power Law \& IID]{
    \centering\includegraphics[width=0.32\linewidth]{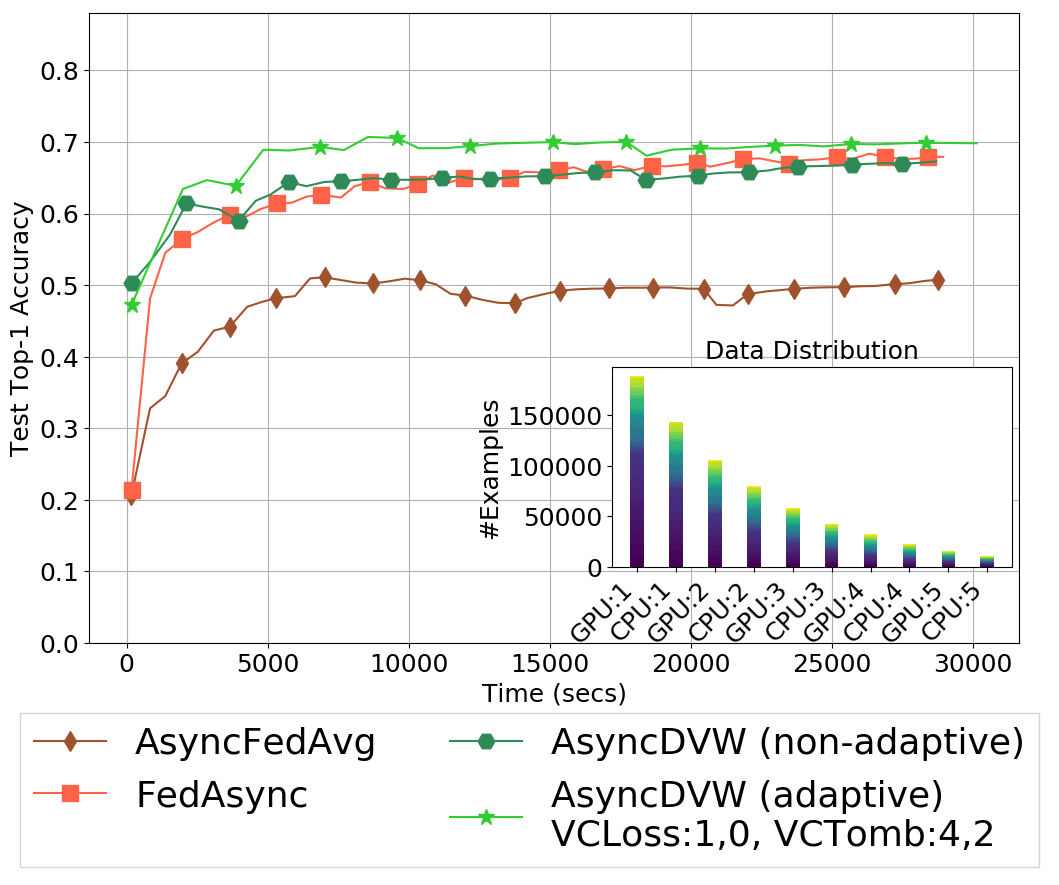}
    \label{subfig:ExtednedMNIST_ByClass_HeterogeneousCluster_PowerLaw_IID}
  }
  
  \captionsetup{justification=centering}
  \caption{Wall-Clock Time Convergence for ExtendedMNIST By Class on a Heterogeneous Cluster}
    \label{fig:ExtendedMNIST_ByClass_AsyncrhonousHeterogeneousCluster}
\end{figure*}

In order to simulate realistic learning environments, we sort each configuration in descending data size order and we assign the data to each learner in an alternating fashion (i.e., fast learner, slow learner, fast learner), except for the uniform distributions where the data size is identical for all learners. 
%
%
Due to space limitations, for every experiment we include the respective data distribution configuration as an inset in the wall-clock time convergence rate plot (Figures~\ref{fig:Cifar10SynchronousHomogeneousCluster}, \ref{fig:Cifar10SynchronousAsyncrhonousHeterogeneousCluster},
\ref{fig:Cifar100AsyncrhonousHeterogeneousCluster}, and
\ref{fig:ExtendedMNIST_ByClass_AsyncrhonousHeterogeneousCluster}).

\begin{figure*}[htpb]
    \centering\includegraphics[width=\linewidth]{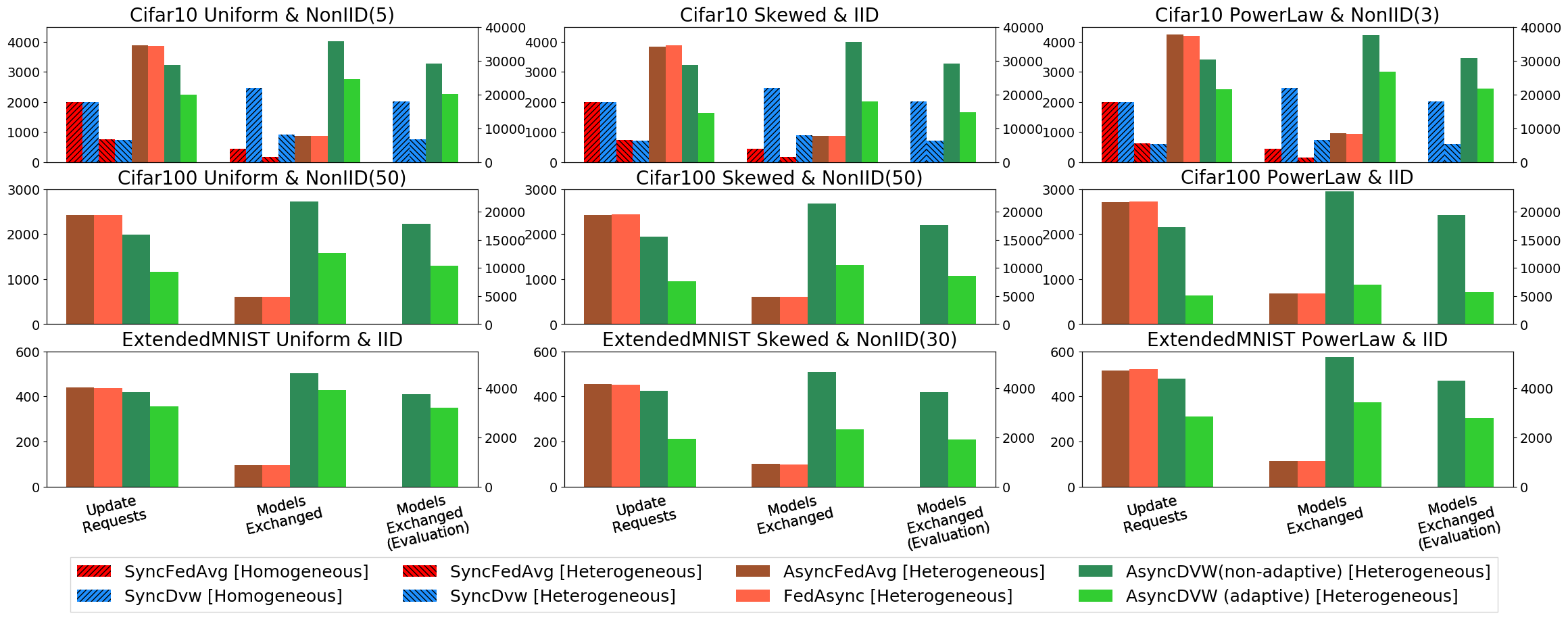}
    \vspace{-5mm} 
    \captionsetup{justification=centering}
    \caption{Training schemes comparison in terms of Total Update Requests (left y-axis), Total Models Exchanged and Total Models Exchanged for DVW Evaluation (right y-axis).}
    \label{fig:FederationStatisticsEvaluation}
\end{figure*}

\textbf{Results.}
We evaluate all the policies on Top-1 accuracy on the separate Test set over wall-clock time. 
%
%
We first evaluate the convergence of the FedAvg and DVW schemes in a homogeneous cluster with synchronous communication in the Cifar10 domain (Figure~\ref{fig:Cifar10SynchronousHomogeneousCluster}). Both SyncFedAvg and SyncDVW were run for 200 communication rounds and the relative difference in wall-clock time is due to the distributed evaluation operations that DVW needs to perform (see model requests in Figure \ref{fig:FederationStatisticsEvaluation}). 
For uniform and moderately skewed data distributions both approaches are comparable. In a synchronous environment, the weighting scheme of FedAvg, which is based on number of examples per learner, is a good proxy to train a federation model. However, as we move towards more severely skewed and imbalanced data distributions, DVW significantly outperforms FedAvg (Figures~\ref{subfig:Cifar10_HomogeneousCluster_PowerLaw_IID}, \ref{subfig:Cifar10_HomogeneousCluster_PowerLaw_NonIID_5}, and \ref{subfig:Cifar10_HomogeneousCluster_PowerLaw_NonIID_3}).

For a homogeneous computing environment (10 GPUs), the left side of Table~\ref{tbl:Cifar10SynchronousAsyncrhonousHeterogeneousCluster} shows the Top-1 accuracy on the Test set after 10,000 seconds of wall-clock time and after 200 communication rounds. The percentage difference in accuracy between  SyncFedAvg and SyncDVW ranges from 1\% to 2\% for uniform and moderately skewed data sizes, but SyncDVW is between 9\% and 27\% better for power law distributions. We do not perform any additional experiments for the homogeneous case, since our main focus is on federated learning over heterogeneous environments.

\begin{table}[htpb]
\noindent
\tiny
\setlength\tabcolsep{0pt}
\captionsetup{justification=centering}
\caption{Cifar10 \\ Homogeneous and Heterogeneous Clusters}
    \centering
    \begin{tabular}{@{}llcccccccccc@{}}
    \cmidrule(l){3-12}
     &
      &
      \multicolumn{4}{c}{\textbf{\begin{tabular}[c]{@{}c@{}}Homogeneous Cluster \\ (x10 GPUs)\end{tabular}}} &
      \multicolumn{6}{c}{\textbf{\begin{tabular}[c]{@{}c@{}}Heterogeneous Cluster\\ (x5 GPUs, x5 CPUs)\end{tabular}}} \\ \cmidrule(lr){3-6} \cmidrule(l ){7-12}
     &
      &
      \multicolumn{2}{c}{\textbf{Acc@10Ksecs}} &
      \multicolumn{2}{c}{\textbf{ Acc@200Rounds}} &
      \multicolumn{6}{c}{\textbf{Acc@14Ksecs}} \\ \cmidrule(lr){3-4} \cmidrule(lr){5-6} \cmidrule(l ){7-12}
    \textbf{\begin{tabular}[c]{@{}l@{}}Data\\ Size\end{tabular}} &
      \textbf{\begin{tabular}[c]{@{}l@{}}Class\\ Distrib.\end{tabular}} &
      \multicolumn{1}{c}{\textbf{\begin{tabular}[c]{@{}c@{}}Sync\\FedAvg\end{tabular}}} &
      \multicolumn{1}{c}{\textbf{\begin{tabular}[c]{@{}c@{}}Sync\\DVW\end{tabular}}} &
      \multicolumn{1}{c}{\textbf{\begin{tabular}[c]{@{}c@{}}Sync\\FedAvg\end{tabular}}} &
      \multicolumn{1}{c}{\textbf{\begin{tabular}[c]{@{}c@{}}Sync\\DVW\end{tabular}}} &
      \multicolumn{1}{c}{\textbf{\begin{tabular}[c]{@{}c@{}}Sync\\FedAvg\end{tabular}}} &
      \multicolumn{1}{c}{\textbf{\begin{tabular}[c]{@{}c@{}}Sync\\DVW\end{tabular}}} &
      \multicolumn{1}{c}{\textbf{\begin{tabular}[c]{@{}c@{}}Async\\FedAvg\end{tabular}}} &
      \multicolumn{1}{c}{\textbf{\begin{tabular}[c]{@{}c@{}}Fed\\Async\end{tabular}}} &
      \multicolumn{1}{c}{\textbf{\begin{tabular}[c]{@{}c@{}}Async\\DVW(na)\end{tabular}}} &
      \multicolumn{1}{c}{\textbf{\begin{tabular}[c]{@{}c@{}}Async\\DVW(a)\end{tabular}}} \\ \toprule
        &
        IID &
          \textbf{0.828} &
          0.825 &
          \textbf{0.8295} &
          0.8285 &
          \textbf{0.8272} &
          0.8207 &
          0.826 &
          0.8045 &
          0.822 & 
          \textbf{0.8272}
      \\
     \textbf{Uniform} &
     Non-IID(5) &
          \textbf{0.8002} &
          0.7968 &
          \textbf{0.8002} &
          0.7994 &
          0.7933 &
          0.7916 &
          0.8009 &
          0.7646 &
          0.7993 &
          \textbf{0.8021}
      \\
     &
     Non-IID(3) &
          \textbf{0.7576} &
          0.7550 &
          \textbf{0.7576} &
          0.7554 &
          0.7461 &
          0.7453 &
          0.7551 &
          0.6897 &
          0.7553 &
          \textbf{0.7559}
      \\ \midrule
    &
     IID &
          \textbf{0.8037} &
          0.8005 &
          0.8054 &
          \textbf{0.8066} &
          0.789 &
          0.7907 & 
          0.7927 &
          0.7594 &
          0.7974 &
          \textbf{0.8072}
      \\
     \textbf{Skewed} &
     Non-IID(5) &
          0.7193 &
          \textbf{0.7288} &
          0.7224 &
          \textbf{0.7372} &
          0.7062 &
          0.7157 &
          0.7262 &
          0.6428 &
          0.7324 &
          \textbf{0.7322}
      \\
     &
     Non-IID(3) &
          0.6852 &
          \textbf{0.6949} &
          0.6868 &
          \textbf{0.6965} &
          0.6681 &
          0.6718 &
          0.6824 &
          0.5564 &
          0.6873 &
          \textbf{0.6943}
      \\ \midrule
     &
     IID &
          0.6998 &
          \textbf{0.7664} & 
          0.7113 & 
          \textbf{0.7709} & 
          0.5149 & 
          0.7221 & 
          0.6507 & 
          0.7248 &
          0.7537 &
          \textbf{0.7682}
      \\
     \textbf{Power Law} &
     Non-IID(5x7) &
          0.6066 &
          \textbf{0.6967} &
          0.6407 &
          \textbf{0.7163} &
          0.3016 &
          0.6281 &
          0.4245 &
          0.5672 &
          0.6807 &
          \textbf{0.6948}
      \\
     &
     Non-IID(3x8) &
          0.4164 &
          \textbf{0.5744} &
          0.4869 &
          \textbf{0.6191} &
          0.1202 &
          0.3546 &
          0.1461 &
          0.4205 &
          0.5517 &
          \textbf{0.5703} \\ \bottomrule
    \end{tabular}
\label{tbl:Cifar10SynchronousAsyncrhonousHeterogeneousCluster}
\end{table}



For a heterogeneous environment, we first compare synchronous and asynchronous policies in the Cifar10 domain. The more heterogeneous the data distributions across the learners, the better the performance of DVW compared to FedAvg and FedAsync, as shown in Figure~\ref{fig:Cifar10SynchronousAsyncrhonousHeterogeneousCluster} and the right side of  Table~\ref{tbl:Cifar10SynchronousAsyncrhonousHeterogeneousCluster}.
For uniform data sizes, FedAvg and DVW perform comparably both in synchronous and asynchronous modes, while FedAsync suffers the more Non-IID the distribution is (\Cref{subfig:Cifar10_HeterogeneousCluster_Uniform_NonIID_5,subfig:Cifar10_HeterogeneousCluster_Uniform_NonIID_3}). 
In moderately skewed domains, AsyncDVW outperforms all other methods, converging initially at a faster rate. FedAsync suffers significantly as the number of classes per learner decreases (\Cref{subfig:Cifar10_HeterogeneousCluster_Skewed_IID,subfig:Cifar10_HeterogeneousCluster_Skewed_NonIID_5,subfig:Cifar10_HeterogeneousCluster_Skewed_NonIID_3}). 
For the Power Law distributions  (\cref{subfig:Cifar10_HeterogeneousCluster_PowerLaw_IID,subfig:Cifar10_HeterogeneousCluster_PowerLaw_NonIID_5,subfig:Cifar10_HeterogeneousCluster_PowerLaw_NonIID_3}),  
AsyncDVW continues to outperform all other methods. In the extreme case of Power Law and Non-IID(3), SyncFedAvg fails to learn within the allocated time, which is consistent with the homogeneous case (Figure~\ref{subfig:Cifar10_HomogeneousCluster_PowerLaw_NonIID_3}) where SyncFedAvg goes through an extended plateau before starting to learn; now the presence of slow learners exacerbates the problem.

Interestingly, FedAsync initially performs better than AsyncDVW in one PowerLaw case (Figure~\ref{subfig:Cifar10_HeterogeneousCluster_PowerLaw_NonIID_3}). 
FedAsync takes into account only the staleness of each learner when computing the community model. Thus, the faster the learner, the higher its contribution to the federation model. As we described earlier, in the Power Law settings, the learners close to the head of the distribution hold data from multiple classes (starting with a fast learner). We hypothesise that the early good performance of FedAsync is driven by these data-rich fast learners. 
The right side of Table~\ref{tbl:Cifar10SynchronousAsyncrhonousHeterogeneousCluster} shows the Test Top-1 accuracy after 14,000 seconds of wall-clock time in the heterogeneous computing environment. AsyncDVW matches or outperforms all other systems.

%

We scaled the experiments to more complex domains, Cifar100 and ExtendedMNIST By Class, showing results for heterogeneous computational environments and asynchronous communication policies in Figures~\ref{fig:Cifar100AsyncrhonousHeterogeneousCluster}, \ref{fig:ExtendedMNIST_ByClass_AsyncrhonousHeterogeneousCluster}, and \cref{tbl:Cifar100AsyncrhonousHeterogeneousCluster,tbl:ExtendedMNIST_ByClass_AsyncrhonousHeterogeneousCluster}.
Overall, the results confirm the previous findings in Cifar10; AsyncDVW outperforms previous approaches, particularly as the heterogeneity among learners increases.

\begin{table}[htpb]
\noindent
\tiny
\setlength\tabcolsep{0pt}
    \begin{minipage}{.45\linewidth}
        \captionsetup{justification=centering}
        \caption{Cifar100 Heterogeneous Cluster}
        \begin{tabular}{@{}llcccc@{}}
            \cmidrule{3-6}
             &  & \multicolumn{4}{c}{\textbf{\begin{tabular}[c]{@{}l@{}}Acc@14Ksecs\end{tabular}}} \\ 
             \cmidrule{3-6}
            \textbf{\begin{tabular}[c]{@{}l@{}}Data\\Size\end{tabular}} &
              \textbf{\begin{tabular}[c]{@{}l@{}}Class\\Distrib.\end{tabular}} &
              \textbf{\begin{tabular}[c]{@{}c@{}}Async\\FedAvg\end{tabular}} &
              \textbf{\begin{tabular}[c]{@{}c@{}}Fed\\Async\end{tabular}} &
              \textbf{\begin{tabular}[c]{@{}c@{}}Async\\DVW(na)\end{tabular}} &
              \textbf{\begin{tabular}[c]{@{}c@{}}Async\\DVW(a)\end{tabular}} \\ \toprule
            \textbf{Uniform}          & IID         &   0.5962   &  0.5745 & 0.5931 & \textbf{0.5996}            \\
                      & Non-IID(50) &   0.5189   & 0.4944 & 0.5179 & \textbf{0.558}      \\ \midrule
            \textbf{Skewed}          & IID         &   0.5437  & 0.5342  & 0.5656 & \textbf{0.5774}               \\
                      & Non-IID(50) &   0.3402    &  0.3525 & 0.3687 & \textbf{0.4459}      \\ \midrule
            \textbf{Power Law}          & IID         &   0.1653 &  0.3512 & 0.4301  & \textbf{0.4444}      \\
                      & Non-IID(50x5) &   0.0193   &  0.186 & 0.2324  & \textbf{0.2605}     \\ \bottomrule
        \end{tabular}
        \label{tbl:Cifar100AsyncrhonousHeterogeneousCluster}
    \end{minipage}%
    \qquad\quad
    \begin{minipage}{.45\linewidth}
        \captionsetup{justification=centering}
        \caption{ExtendedMNIST Heterogeneous Cluster}
        \begin{tabular}{@{}llcccc@{}}
            \cmidrule{3-6}
             & & \multicolumn{4}{c}{\textbf{\begin{tabular}[c]{@{}l@{}}Acc@30Ksecs\end{tabular}}} \\ \cmidrule{3-6}
            \textbf{\begin{tabular}[c]{@{}l@{}}Data \\ Size\end{tabular}} &
              \textbf{\begin{tabular}[c]{@{}l@{}}Class \\ Distrib.\end{tabular}} &
              \textbf{\begin{tabular}[c]{@{}c@{}}Async\\ FedAvg\end{tabular}} &
              \textbf{\begin{tabular}[c]{@{}c@{}}Fed\\Async\end{tabular}} &
              \textbf{\begin{tabular}[c]{@{}c@{}}Async\\DVW(na)\end{tabular}} &
              \textbf{\begin{tabular}[c]{@{}c@{}}Async\\DVW(a)\end{tabular}} \\ \toprule
                     \textbf{Uniform} & IID         &   0.8586     & 0.86 & 0.8588 &  \textbf{0.8631}            \\ \midrule
            \textbf{Skewed}          & Non-IID(30)         &   0.7911  & 0.7346 & 0.8103 & \textbf{0.8285}               \\ \midrule
            \textbf{Power Law}          & IID         &   0.5107  &  0.6833 & 0.6723 & \textbf{0.7107}      \\ \bottomrule
        \end{tabular}
        \label{tbl:ExtendedMNIST_ByClass_AsyncrhonousHeterogeneousCluster}
    \end{minipage}
\end{table}

In the ExtendedMNIST domain, we observe oscillations in the performance of the federation across most systems, but AsyncFedAvg is the most affected. Since this domain contains many more examples and takes longer to train, the impact of the slow learners is very significant for AsyncFedAvg. AsyncDVW and FedAsync are more resilient to these fluctuations given that their weight mixing strategies take into account the performance of the learners, either directly as in DVW, or indirectly through model staleness in FedAsync. We believe we can reduce these oscillations by tuning the hyperparameter values during training, which is part of our future work. 

Overall, DVW is more robust across different domains and diverse data distributions than other weighting schemes.
We attribute the faster convergence of adaptive AsyncDVW at the beginning of the training phase in most of the experiments to our update frequency criteria. The adaptivity of the update frequency of a learner often results in performing more local iterations at the beginning of the federation, which boosts overall performance. By assigning a good estimate of the value of a local model, the DVW approach makes learning more resilient to diverse environments.

Figure \ref{fig:FederationStatisticsEvaluation} shows communication costs in the federation across all the policies and domains (one experimental setup per data size). We compare the policies in terms of total number of update requests (leftmost column group, left y-axis) and total number of models exchanged (central group, right y-axis). For DVW we also show the difference in models exchanged for evaluation (rightmost group, right y-axis). For all policies except DVW the number of models exchanged is equal to $Update Requests \times 2$ (i.e., send 1 local model to controller, receive 1 community model from controller) and for DVW it is $Update Requests \times 11$ (i.e., requesting learner sends 1 local model to controller along with its local validation score, controller sends the local model to each evaluation service (x9) and receives scalar values, and requesting learner receives 1 community model from controller). Overall, DVW and especially adaptive DVW perform fewer community requests compared to other asynchronous approaches. 
The total number of models exchanged during training required by DVW is greater than other approaches, due to the additional model evaluations. 
Adaptive DVW requires a smaller number of model evaluations and smaller number of exchanged models than Non-Adaptive DVW, performing comparably to other asynchronous policies.  
%
In general, the additional model exchange cost of DVW trades off for better generalization.

\section{Discussion}\label{sec:Discussion}

We presented a modular architecture for Federated Learning in environments with heterogeneous computational resources and data distributions. We developed a novel federation aggregation scheme,  Distributed Validation Weighting, that seeks to measure directly the quality of each learner in a federation by evaluating its local model over a distributed validation set. The DVW method performs well across a wide variety of target class data distributions and number of examples per learner. 

Even though our Asynchronous DVW protocol is trained over a smaller dataset (i.e., 95\% of the training data, since 5\% of each learner training data is reserved for validation), it can learn significantly faster and train more robust federation models than state-of-the-art methods, such as Synchronous and Asynchronous FedAvg, and FedAsync, which use all of the training data.

Our empirical results over a range of data distributions in challenging domains (e.g., Cifar-100, ExtendedMNIST By Class) and complex networks (e.g., ResNet-50) demonstrate that our Asynchronous DVW  scheme is well-suited for heterogeneous environments with diverse computational resources and data distributions. We expect these types of environments to be common in medical and industrial applications, such as a consortium of hospitals that wants to analyze medical records without pooling the data on a centralized location.


Our immediate future work includes investigating methods to leverage the complete training data in a post-validation phase, and sampling learners based on their local data distribution when generating the distributed validation set. We are also investigating protocols that perform online hyperparameter tuning, such as adaptive learning rate decay using distributed validation loss as an evaluation metric. Finally, we plan to evaluate the DVW scheme on additional classification and regression tasks. 


\bibliographystyle{IEEEtran}
\bibliography{arxiv_submission.bib} 

\end{document}